\documentclass[10pt,twocolumn,letterpaper]{article}

\usepackage{cvpr}
\usepackage{times}
\usepackage{epsfig}
\usepackage{graphicx}
\usepackage{amsmath}
\usepackage{amssymb}
\usepackage{color}
\usepackage{subfig}
\usepackage{multirow}

\usepackage[pagebackref=true,breaklinks=true,letterpaper=true,colorlinks,bookmarks=false]{hyperref}

\usepackage{appendix}
\usepackage{siunitx}
\usepackage{amsthm}
\newtheorem{definition}{Definition}
\usepackage{listings}
\usepackage{threeparttable}
\usepackage{booktabs}
\usepackage{xcolor}
\usepackage{diagbox}
\usepackage{enumerate}
\usepackage{placeins}
\usepackage{algorithm}
\usepackage{algorithmic}

\definecolor{mygreen}{rgb}{0,0.6,0}
\definecolor{myblue}{rgb}{0,0,0.8}
\definecolor{myorange}{rgb}{0.6,0.4,0.2}
\definecolor{myblack}{rgb}{0,0,0}

\makeatletter

\newcommand{\Rmnum}[1]{\expandafter\@slowromancap\romannumeral #1@}
\makeatother

\cvprfinalcopy

\def\cvprPaperID{0000}

\providecommand{\IEEEPARstart}[2]{#1#2}
\providecommand{\appendices}{\appendix}
\newcommand{\citep}[1]{\cite{#1}}
\newlength{\mathindent}

\ifcvprfinal\fi

\begin{document}

\title{Compound and Parallel Modes of Tropical Convolutional Neural Networks}

\author{Mingbo Li \thanks{Equal contribution}
\qquad
Liying Liu \footnotemark[1] 
\qquad
Charles Wiranto
\qquad
Ye Luo \thanks{Corresponding author} \\ 
\vspace{0.2em}\\
Xiamen University\\
{\tt\small \{limingbo,llyxmu185,charleswiranto\}@stu.xmu.edu.cn}, \tt\small \{luoye\}@xmu.edu.cn
}

\maketitle

\begin{abstract}
Convolutional neural networks (CNNs) are foundational to many state-of-the-art computer vision systems, yet their reliance on multiplication-intensive computations poses challenges for deployment on resource-constrained devices. While tropical convolutional neural networks (TCNNs) reduce this computational burden by replacing multiplications with cheaper min/max-plus operations, they often do so at the cost of reduced model accuracy. To address this trade-off, we introduce two novel extensions of tropical convolution: compound tropical convolution (cTCNN) and parallel tropical convolution (pTCNN). These operators combine min-plus and max-plus algebraic operations within a single layer to enhance representational capacity while maintaining low computational cost. We provide an open-source implementation of these operators in a PyTorch-compatible framework, featuring optimized GPU kernels developed with TileLang. Through extensive experiments on image classification and semantic segmentation benchmarks, we demonstrate that our proposed cTCNN and pTCNN layers achieve competitive performance against standard CNNs while significantly reducing the number of multiplications. Moreover, we show that hybrid models, which integrate both tropical and conventional convolutions, can further improve the accuracy-efficiency balance. Our findings suggest that these tropical convolution variants are viable and effective components for building efficient deep learning models.
\end{abstract}

\section{Introduction}\label{sec:intro}
\IEEEPARstart{T}{he} remarkable feature extraction capabilities of convolutional neural networks (CNNs) have made them a key driving force in the advancement of machine learning. In recent years, deep learning methods based on CNNs have been successfully applied to various tasks, including speech recognition~\citep{abdel-hamid_convolutional_2014,sainath_convolutional_2015,dai2017very}, image classification~\citep{he_deep_2016,huang_densely_2017}, object detection~\citep{bochkovskiy_yolov4_2020,li_yolov6_2022,wang_yolov7_2022}, and background subtraction~\citep{tezcan_bsuv-net_2020,babaee_deep_2018,lim_foreground_2018,lim_learning_2020}. Despite their superior performance, these CNNs typically require high-power GPUs for training and inference due to the extensive multiplication operations involved, which limits their large-scale deployment, particularly on edge devices such as smartphones and smartwatches in the future.

To address this limitation, a range of neural network compression techniques have been developed in recent years, some of which aim to reduce the number of multiplications in network calculations without significantly compromising performance. One of the pioneering works in this area, BinaryConnect~\citep{courbariaux_binaryconnect_2015}, enforces the network weights, typically represented as floating-point values, to be binary (-1 or 1). This transforms multiplication operations into simple addition and subtraction, thereby compressing the model size and accelerating computations. ~\cite{chen_addernet_2020} introduced AdderNet, which replaces the traditional cross-correlation operation with $L_1$ distance to eliminate multiplications, drawing on the similarity between the image and the convolution kernel. Subsequently, ~\cite{fan2021alternative} proposed TCNN, which utilizes a novel convolution operation based on tropical geometry to further reduce multiplications. However, these approaches, while effective in minimizing multiplication operations, often lead to overly simplistic network structures that compromise feature extraction capabilities. Consequently, they struggle to balance computational complexity with model performance, falling short of achieving results comparable to traditional CNNs.

In this paper, we propose leveraging tropical algebra to extend convolutional neural networks (CNNs). Building on the previously introduced tropical convolutional neural networks (TCNN), we introduce two more generalized forms: compound tropical convolutional neural networks (cTCNN) and parallel tropical convolutional neural networks (pTCNN). These convolution operators, grounded in tropical geometry, extend TCNN's design and aim to better balance computational cost and model capacity.


The cTCNN combines the functions of two TCNN layers using the same weight, allowing for both min-plus and max-plus operations within the same convolutional layer. Compared to traditional CNNs, cTCNN benefits from TCNN's ability to maximize addition and minimize multiplication, thereby reducing model complexity. Furthermore, unlike TCNN, the sliding window in each cTCNN convolution operation incorporates feature information from both min-plus and max-plus operations, leading to more diverse feature extraction and improved model performance. In contrast, pTCNN combines the functions of two TCNN layers using different weights, offering more adjustable parameters and enabling even more diverse feature extraction, further enhancing model performance.

We evaluate the proposed cTCNN/pTCNN operators on a range of computer vision benchmarks. The main paper reports two-dimensional image classification experiments and semantic segmentation experiments, and investigates how these operators behave when integrated into deeper backbones (e.g., ResNet and MobileNetV2). Additional one-dimensional and three-dimensional LeNet-based experiments are provided in Appendix~\ref{sec:1dc-TCNN} and~\ref{sec:3dc-TCNN}. Overall, the results indicate that cTCNN and pTCNN can provide a favorable trade-off between representational capability and computational cost, and can achieve competitive performance compared with conventional CNN baselines under standard training settings.


Our contributions are summarized as follows:

 \begin{enumerate}
\item{
Based on tropical algebra theory, we introduced compound convolutional neural networks (cTCNN) and parallel convolutional neural networks (pTCNN), which represent different combinations of min-plus and max-plus operations.
}

\item{
We designed and implemented a convolutional neural network framework based on tropical algebra, which not only integrates previous works on tropical CNNs but also includes cTCNN and pTCNN. This framework is built on TileLang~\citep{wang2025tilelangcomposabletiledprogramming} and the PyTorch library, allowing it to replace convolutional components in existing PyTorch implementations. The framework is also extensible to support additional tropical and other operations in neural networks. We have made our implementation publicly available on GitHub \footnote{ cTCNN: \url{***}}.
}

\item{
We investigated the integration of cTCNN and pTCNN into deep neural network architectures. Experimental results show that combining these operators with conventional CNN layers can achieve competitive performance compared with standalone CNN baselines, highlighting their complementary feature extraction capabilities.
}

\item{
We evaluated cTCNN and pTCNN on multiple benchmark datasets for image classification and semantic segmentation. The results indicate that the proposed operators can provide a favorable trade-off between representational capability and computational cost while achieving competitive performance across different tasks.
}

\item{
We further extended tropical convolutional operators to one-dimensional and three-dimensional tasks using LeNet-style architectures. Additional experimental results are provided in the appendix.

}
\end{enumerate}

The remainder of this paper is organized as follows. Section~\ref{sec:related} reviews related work. Section~\ref{sec:method} presents the proposed tropical convolution operators and network constructions. Section~\ref{sec:analysis} provides theoretical and computational analyses. Section~\ref{sec:implementation} describes the implementation details of our framework. Section~\ref{sec:experiments} reports experimental results on image classification and semantic segmentation tasks. Finally, Section~\ref{sec:conclusion} concludes the paper and discusses future work.

\section{Related work}\label{sec:related}

Tropical algebra~\citep{maclagan2009introduction} provides algebraic structures in which conventional addition and multiplication are replaced by min/max and addition operations. It has been used in several areas of mathematics and computer science. One representative application is network calculus~\citep{le2001network}, where min-plus algebra is used to analyze service curves, delay, backlog, and loss bounds in networked systems~\citep{le1998network,jiang2008stochastic,liebeherrmin}. In neural-network research, tropical algebra has also been used as an analysis tool. For example, \cite{zhang2018tropical} studied the complexity of feedforward neural networks with ReLU activations, \cite{alfarra2020decision} described neural-network decision boundaries from a tropical-geometry perspective, and related connections between ReLU networks and tropical polyhedra have been studied in~\citep{fournier2019tropical,goubault2021static}. More recently, \cite{Pasque2024TropicalDB} proposed a tropical CNN based on embeddings into tropical projective tori for adversarial robustness. These studies suggest that tropical algebra can provide both analytical tools and alternative computational primitives for neural networks.

Reducing the cost of neural-network computation is a broad research direction, and existing methods address different sources of cost. One line of work reduces numerical precision. BinaryConnect~\citep{courbariaux2015binaryconnect} constrains weights to binary values, BNNs~\citep{hubara2016binarized} also binarize activations, and XNOR-Net~\citep{rastegari2016xnor} introduces scaling factors to reduce the loss caused by binarization. ABC-Net~\citep{lin2017towards} further uses linear combinations of binary bases to improve the approximation of full-precision weights. These methods can reduce memory and arithmetic cost, but they mainly rely on quantization and therefore face the usual trade-off between numerical precision and accuracy.

Another related direction is lightweight architecture design. Networks such as SqueezeNet~\citep{gholami2018squeezenext,iandola2016squeezenet}, ShuffleNet~\citep{ma2018shufflenet,zhang2018shufflenet}, CondenseNet~\citep{yang2021condensenet,huang2018condensenet}, MobileNet~\citep{howard2019searching,howard2017mobilenets,sandler2018mobilenetv2,zhou2020rethinking}, and EfficientNet~\citep{tan2019efficientnet,tan2021efficientnetv2} reduce parameters and floating-point operations through carefully designed convolutional blocks, channel grouping, depthwise separable convolution, or compound scaling. These approaches are highly relevant baselines for efficient vision models. However, their main focus is architectural efficiency, while our work focuses on the arithmetic form of the convolution operator itself. The two directions are not mutually exclusive; in this paper we therefore also study tropical operators inside deeper backbones such as ResNet and MobileNetV2.

Several methods are closer to our work because they replace multiplication-heavy operations with cheaper arithmetic primitives. AdderNet~\citep{chen2020addernet} replaces convolutional correlation with an $\ell_1$-distance-based operation, thereby avoiding multiplications inside the convolution operator. ShiftAddNet~\citep{you2020shiftaddnet} and ShiftAddViT~\citep{you2024shiftaddvit} use shift and addition operations to approximate multiplication-heavy computation. These methods show that changing the operator-level arithmetic can be useful, but the resulting accuracy and speed depend on optimization behavior and hardware support. Our work follows the same broad operator-level motivation, but uses min-plus and max-plus operations from tropical algebra rather than distance-based or shift-add approximations.

Recent efficient Transformer and sequence-model studies address a different but relevant efficiency problem. Starting from the Transformer~\citep{vaswani2017attention}, methods such as Mamba~\citep{gu2023mamba}, Mamba-2~\citep{dao2024transformers}, RWKV~\citep{peng2023rwkv}, RetNet~\citep{sun2023retentive}, Hyena~\citep{poli2023hyena}, simplified Transformer blocks~\citep{he2023simplifying}, and LookupViT~\citep{koner2024lookupvit} improve sequence modeling or token processing through alternative mixing mechanisms, state-space formulations, recurrent-style designs, or token selection. These works are important for efficient attention and sequence pipelines. In contrast, the present paper studies convolutional operators in CNN-style vision backbones. Therefore, we do not claim that tropical convolution replaces efficient Transformer designs; rather, it addresses a complementary question: how to reduce multiplication-heavy computation inside convolutional layers while retaining useful representational capacity.

Tropical-algebra-based neural-network operators are most directly related to this paper. \cite{luo2021min} proposed min-max-plus neural networks and analyzed their approximation ability with limited multiplications. \cite{fan2021alternative} introduced tropical convolutional neural networks (TCNNs), which replace the multiply--accumulate pattern of standard convolution with addition and min/max operations. Such operators are also relevant to resource-constrained settings such as TinyML~\citep{lin2023tiny}. However, directly replacing a standard convolution with a single min-plus or max-plus tropical convolution can limit the information captured by each receptive field. Motivated by this limitation, we investigate compound and parallel combinations of min-plus and max-plus convolutions. The goal is modest: to study whether combining the two tropical responses can improve the accuracy-efficiency trade-off of TCNN-style layers, and to evaluate this behavior in both controlled lightweight models and deeper CNN backbones.

\section{Method}\label{sec:method}
In this section, we first briefly review tropical algebra and tropical convolution, and then extend the tropical convolutional neural network (TCNN) proposed by Fan et al.~\citep{fan2021alternative} from the two-dimensional setting to one-dimensional and three-dimensional cases. Next, we introduce the mathematical formulation of compound tropical convolution, describe the implementation of compound tropical convolutional layers in different dimensions, and finally extend compound tropical convolution to parallel tropical convolution.

\subsection {Tropical algebra and tropical convolution}\label{subsec:tcnn}
Tropical algebra includes the min-plus and max-plus algebras with min and max being dual operations. More specifically, min-plus algebra is defined as $\mathbb{R}_{\min}=\{\mathbb{R}\cup\{\infty \},\oplus,\odot\}$ where $x \oplus y:=\min(x,y)$ and $x \odot y:=x+y$ for all  $x,y\in \mathbb{R}_{\min}$, and max-plus algebra is defined as $\mathbb{R}_{\max}=\{\mathbb{R}\cup\{-\infty \},\boxplus,\odot\}$
	where $x \boxplus y:=\max(x,y)$ and $x \odot y:=x+y$ for all $x,y\in \mathbb{R}_{\max}$. Note that $\infty$ and $-\infty$ are additive identity elements of the min-plus and max-plus algebras, respectively, which are sometimes omitted when computations only involve finite numbers.
    
Next we compare the notion of traditional convolution and the notion of convolution in the sense of tropical algebra.

The (one-dimensional) convolution of two functions $f$ and $g$ in the traditional sense can be expressed as
$(f * g)(t) = \int_{0}^{t} f(\tau) g(t-\tau) d \tau$ where $t$ can be defined on any domain, e.g., the time domain. Variants of this
convolution operation are widely used in deep learning, and a large number of multiplication operations will appear in model training, bringing challenges to hardware devices. 

In tropical algebra, addition is replaced by taking the larger/smaller values, and multiplication is replaced by addition.  Hence, one can construct analogously a convolution operation using tropical algebra. Specific definitions are provided as follows:
\begin{definition} 
Tropical convolutions: 
\begin{enumerate}
	\item[(i)] The \emph{(tropical) min-plus convolution} of two function $f$ and $g$ is defined as:
	\begin{flalign} 
        &(f \underline{\otimes} g)(t)=\inf\limits_{t \ge \tau \ge 0}(f(\tau)+g(t-\tau))&
    \end{flalign}
	\item[(ii)] The \emph{(tropical) max-plus convolution} of two function $f$ and $g$ is defined as:
    \begin{flalign} 
        &(f \overline{\otimes} g)(t)=\sup\limits_{t \ge \tau \ge 0}(f(\tau)+g(t-\tau))&
    \end{flalign}
\end{enumerate} 
\end{definition}

In ~\cite{fan2021alternative}, this convolution operation is used to replace the traditional convolution operation in CNNs to construct tropical convolutional neural networks, or TCNNs, which are applied to image classification tasks with good performance. 
However, their work was limited to two-dimensional tasks. In this paper, we considered how to extend the tropical convolution to one-dimensional and three-dimensional tasks, as well as how to combine the two types of tropical convolution operations to enhance the model's performance across different dimensional tasks.
Firstly, let us recall the definition of tropical convolution in two-dimensional tasks in ~\cite{fan2021alternative}.
The weight of a two-dimensional convolutional kernel can be represented by $W \in \mathbb{R}^{K_{\text{h}} \times K_{\text{w}} \times C_{\text{in}} \times C_{\text{out}}}$, where $C_{\text{in}}$ is the input channel size and $K_{\text{h}} \times K_{\text{w}}$ is the kernel size. The input tensor can be represented by $X \in \mathbb{R}^{H_{\text{in}} \times W_{\text{in}} \times C_{\text{in}}}$, where $H_{\text{in}}$ and $W_{\text{in}}$ are the height and width of the input tensor, respectively. And the output characteristic $Y \in \mathbb{R}^{H_{\text{out}} \times W_{\text{out}} \times C_{\text{out}}}$ can be calculated as the tropical convolution of $X$ and $W$.\par

Let $h$, $w$,  and $c_{\text{out}}$ be the indices of the output tensor. In particular, the index  $c_{\text{out}}$ refers to the $c_{\text{out}}$-th channel of the output tensor. Then, we can formulate six TCNN operations for the two-dimensional case as follows:

First, let's define the base operator.
\begin{flalign}
    \text{MinPlus2d}&(X,W, h, w, c_{\text{in}}, c_{\text{out}}) \nonumber\\
    &=\min_{(i,j) \in [K_{\text{h}}]\times [K_{\text{w}}]} 
    \Big(X_{h+i, w+j, c_{\text{in}}}+ W_{i,j,c_{\text{in}},c_{\text{out}}}\Big)
\end{flalign}
and 
\begin{flalign}
    \text{MaxPlus2d}&(X,W, h, w, c_{\text{in}}, c_{\text{out}}) \nonumber \\
    &=\max_{(i,j) \in [K_{\text{h}}]\times [K_{\text{w}}]}  
    \Big(X_{h+i, w+j, c_{\text{in}}}+ W_{i,j,c_{\text{in}},c_{\text{out}}}\Big)
\end{flalign}
Then we have the computing formulas for the following convolution layers.
\begin{enumerate}
	\item[(i)] \emph{MinPlus-Sum-Conv2d Layer}:\\ 
	\begin{flalign}
    	Y_{h,w,c_{\text{out}}} =\sum_{c_{\text{in}}\in [C_{\text{in}}]} \text{MinPlus2d}(X,W, h, w, c_{\text{in}}, c_{\text{out}})
    \end{flalign}
	\item[(ii)] \emph{MaxPlus-Sum-Conv2d Layer}: \\
	\begin{flalign}
    	Y_{h,w,c_{\text{out}}} =\sum_{c_{\text{in}}\in [C_{\text{in}}]}\text{MaxPlus2d}(X,W, h, w, c_{\text{in}}, c_{\text{out}})
	\end{flalign} 
	\item[(iii)] \emph{MinPlus-Max-Conv2d Layer}: \\
   	\begin{flalign}
    	Y_{h,w,c_{\text{out}}} =\max_{c_{\text{in}}\in [C_{\text{in}}]}\text{MinPlus2d}(X,W, h, w, c_{\text{in}}, c_{\text{out}})
	\end{flalign}
	\item[(iv)] \emph{MinPlus-Min-Conv2d Layer}: \\
	\begin{flalign}
    	Y_{h,w,c_{\text{out}}} =\min_{c_{\text{in}}\in [C_{\text{in}}]}\text{MinPlus2d}(X,W, h, w, c_{\text{in}}, c_{\text{out}})
    \end{flalign}
   	\item[(v)] \emph{MaxPlus-Max-Conv2d Layer}:\\
	\begin{flalign}
    	Y_{h,w,c_{\text{out}}} =\max_{c_{\text{in}}\in [C_{\text{in}}]}\text{MaxPlus2d}(X,W, h, w, c_{\text{in}}, c_{\text{out}})
	\end{flalign}
	\item[(vi)] \emph{MaxPlus-Min-Conv2d Layer}:\\
	\begin{flalign}
    	Y_{h,w,c_{\text{out}}} =\min_{c_{\text{in}}\in [C_{\text{in}}]}\text{MaxPlus2d}(X,W, h, w, c_{\text{in}}, c_{\text{out}})
    \end{flalign}
\end{enumerate}


\subsection{Extensions of two-dimensional case} \label{subsec:tcnn1}
We extend the formulation of tropical convolutions to one-dimensional and three-dimensional cases, beyond the two-dimensional scenario. This extension allows the tropical convolutional neural network to be applied to a broader range of tasks and enhances its performance across various applications.

Let us consider the one-dimensional case first.
The weight of a one-dimensional convolutional kernel can be represented by $W \in \mathbb{R}^{K_{\text{l}} \times C_{\text{in}} \times C_{\text{out}}}$, where $C_{\text{in}}$ is the input channel size and $K_{\text{l}}$ is the kernel length. The input tensor can be represented by $X \in \mathbb{R}^{L_{\text{in}} \times C_{\text{in}}}$, where $L_{\text{in}}$ is the length of the input tensor. The output tensor $Y \in \mathbb{R}^{L_{\text{out}} \times C_{\text{out}}}$ is calculated as the tropical convolution of $X$ and $W$.\par

Let $l$ and $c_{\text{out}}$ be indices of the output tensor, where $c_{\text{out}}$ represents the index of the output channel. We can formulate the basic operators for one-dimensional tropical convolution as follows:
\begin{flalign}
    &\text{MinPlus1d}(X,W, l, c_{\text{in}}, c_{\text{out}}) = \min_{i \in [K_{\text{l}}]} 
    \Big(X_{l+i, c_{\text{in}}}+ W_{i,c_{\text{in}},c_{\text{out}}}\Big)
\end{flalign}
and
\begin{flalign}
    &\text{MaxPlus1d}(X,W, l, c_{\text{in}}, c_{\text{out}}) = \max_{i \in [K_{\text{l}}]} 
    \Big(X_{l+i, c_{\text{in}}}+ W_{i,c_{\text{in}},c_{\text{out}}}\Big)
\end{flalign}
Then we have the following  formulas:
\begin{enumerate}
	\item[(i)] \emph{MinPlus-Sum-Conv1d Layer}:\\ 
	\begin{flalign}
    	Y_{l,c_{\text{out}}}=\sum_{c_{\text{in}}\in [C_{\text{in}}]}\text{MinPlus1d}(X,W, l, c_{\text{in}}, c_{\text{out}})
	\end{flalign}
	\item[(ii)] \emph{MaxPlus-Sum-Conv1d Layer}: \\
	\begin{flalign}
    	Y_{l,c_{\text{out}}}=\sum_{c_{\text{in}}\in [C_{\text{in}}]}\text{MaxPlus1d}(X,W, l, c_{\text{in}}, c_{\text{out}})
	\end{flalign}
	\item[(iii)] \emph{MinPlus-Max-Conv1d Layer}: \\
   	\begin{flalign}
    	Y_{l,c_{\text{out}}}=\max_{c_{\text{in}}\in [C_{\text{in}}]}\text{MinPlus1d}(X,W, l, c_{\text{in}}, c_{\text{out}})
	\end{flalign}
	\item[(iv)] \emph{MinPlus-Min-Conv1d Layer}: \\
	\begin{flalign}
		Y_{l,c_{\text{out}}}=\min_{c_{\text{in}}\in [C_{\text{in}}]}\text{MinPlus1d}(X,W, l, c_{\text{in}}, c_{\text{out}})
	\end{flalign}
   	\item[(v)] \emph{MaxPlus-Max-Conv1d Layer}:\\
	\begin{flalign}
    	Y_{l,c_{\text{out}}}=\max_{c_{\text{in}}\in [C_{\text{in}}]}\text{MaxPlus1d}(X,W, l, c_{\text{in}}, c_{\text{out}})
	\end{flalign}
	\item[(vi)] \emph{MaxPlus-Min-Conv1d Layer}:\\
	\begin{flalign}
    	Y_{l,c_{\text{out}}}=\min_{c_{\text{in}}\in [C_{\text{in}}]}\text{MaxPlus1d}(X,W, l, c_{\text{in}}, c_{\text{out}})
	\end{flalign}
\end{enumerate}


Now let us consider the three-dimensional case.
The weight of a three-dimensional convolutional kernel can be represented by $W \in \mathbb{R}^{K_{\text{h}} \times K_{\text{w}} \times K_{\text{d}} \times C_{\text{in}} \times C_{\text{out}}}$, where $C_{\text{in}}$ is the input channel size and $K_{\text{h}} \times K_{\text{w}} \times K_{\text{d}}$ is the kernel size. The input tensor can be represented by $X \in \mathbb{R}^{H_{\text{in}} \times W_{\text{in}} \times D_{\text{in}} \times C_{\text{in}}}$, where $H_{\text{in}}$, $W_{\text{in}}$ and $D_{\text{in}}$ are the height, width and depth of the input tensor, respectively. And the output tensor $Y \in \mathbb{R}^{H_{\text{out}} \times W_{\text{out}} \times D_{\text{out}} \times C_{\text{out}}}$ can be calculated as the tropical convolution of $X$ and $W$.\par

Let $h$, $w$, $d$ and $c_{\text{out}}$ be indices of the output tensor. Then we can formulate the base operators for three-dimensional tropical convolution as follows:
\begin{flalign}
    &\text{MinPlus3d}(X,W, h, w, d, c_{\text{in}}, c_{\text{out}}) \nonumber \\ 
    &=\min_{(i,j,k) \in [K_{\text{h}}]\times [K_{\text{w}}]\times [K_{\text{d}}]} 
    \Big(X_{h+i, w+j, d+k, c_{\text{in}}} + W_{i,j,k,c_{\text{in}},c_{\text{out}}}\Big)
\end{flalign}
and
\begin{flalign}
    &\text{MaxPlus3d}(X,W, h, w, d, c_{\text{in}}, c_{\text{out}}) \nonumber \\
    &= \max_{(i,j,k) \in [K_{\text{h}}]\times [K_{\text{w}}]\times [K_{\text{d}}]} 
    \Big(X_{h+i, w+j, d+k, c_{\text{in}}} + W_{i,j,k,c_{\text{in}},c_{\text{out}}}\Big)
\end{flalign}
Then we can give the formula
\begin{enumerate}
	\item[(i)] \emph{MinPlus-Sum-Conv3d Layer}:\\ 
	\begin{flalign}
    	Y_{h,w,d,c_{\text{out}}}=\sum_{c_{\text{in}}\in [C_{\text{in}}]}\text{MinPlus3d}(X,W, h, w, d, c_{\text{in}}, c_{\text{out}})
    \end{flalign}
	\item[(ii)] \emph{MaxPlus-Sum-Conv3d Layer}: \\
	\begin{flalign}
    	Y_{h,w,d,c_{\text{out}}}=\sum_{c_{\text{in}}\in [C_{\text{in}}]}\text{MaxPlus3d}(X,W, h, w, d, c_{\text{in}}, c_{\text{out}})
	\end{flalign}
	\item[(iii)] \emph{MinPlus-Max-Conv3d Layer}: \\
   	\begin{flalign}
    	Y_{h,w,d,c_{\text{out}}}=\max_{c_{\text{in}}\in [C_{\text{in}}]}\text{MinPlus3d}(X,W, h, w, d, c_{\text{in}}, c_{\text{out}})
    \end{flalign}
   	\item[(iv)] \emph{MinPlus-Min-Conv3d Layer}: \\
   	\begin{flalign}
    	Y_{h,w,d,c_{\text{out}}}=\min_{c_{\text{in}}\in [C_{\text{in}}]}\text{MinPlus3d}(X,W, h, w, d, c_{\text{in}}, c_{\text{out}})
	\end{flalign}
   	\item[(v)] \emph{MaxPlus-Max-Conv3d Layer}:\\
   	\begin{flalign}
    	Y_{h,w,d,c_{\text{out}}}=\max_{c_{\text{in}}\in [C_{\text{in}}]}\text{MaxPlus3d}(X,W, h, w, d, c_{\text{in}}, c_{\text{out}})
    \end{flalign}
	\item[(vi)] \emph{MaxPlus-Min-Conv3d Layer}:\\
	\begin{flalign}
    	Y_{h,w,d,c_{\text{out}}}=\min_{c_{\text{in}}\in [C_{\text{in}}]}\text{MaxPlus3d}(X,W, h, w, d, c_{\text{in}}, c_{\text{out}})
    \end{flalign}
\end{enumerate}


\subsection{Compound and parallel tropical convolutions} \label{new compoud convolutional}
In the TCNNs discussed above, each convolutional kernel relies on either the min-plus or max-plus convolution operation, which can lead to limitations. For instance, min-plus operations may perform better on some datasets, while max-plus operations might be more effective on others, indicating that the features extracted by these operations can differ noticeably. To address this and enhance the flexibility of feature extraction in convolutional layers, we introduce two new convolution operations: the compound and parallel operations. These operations are linear combinations of min-plus and max-plus operations applied within the same kernel, offering a more versatile approach to feature extraction.

Let us first define the compound tropical convolution operation.

\begin{definition}\label{D:compound}
With two additional parameters $\alpha, \beta$ to the convolution operation, we define the \emph{compound tropical convolution with two parameters}:
\begin{flalign}
&(f \ \underline{\overline {\otimes}}\ (g,\alpha, \beta))(t) \nonumber \\
&=\alpha\cdot (f\underline{\otimes} g)(t)+\beta\cdot (f\overline{\otimes} g)(t) \nonumber \\
& = \alpha \cdot \inf\limits_{t \ge \tau \ge 0} (f(\tau) + g(t-\tau)) + \nonumber \\ 
&\quad \beta\cdot \sup\limits_{t \ge \tau \ge 0} (f(\tau)+g(t-\tau))
\end{flalign}

\end{definition}
We can then go on to define the parallel tropical convolution operation:
\begin{definition}\label{D:parallel}
With two additional parameters $\alpha, \beta$ to the convolution operation, we define the \emph{parallel tropical convolution  with two parameters}:\par
\begin{flalign}
&(f \ \underline{\overline {\otimes}}\ (g_1,g_2,\alpha, \beta))(t) \nonumber \\
&=\alpha\cdot (f\underline{\otimes} g_{1})(t)+\beta\cdot (f\overline{\otimes} g_{2})(t) \nonumber \\
& = \alpha \cdot\inf\limits_{t \ge \tau \ge 0} (f(\tau) + g_{1}(t-\tau)) + \nonumber \\ 
&\quad \beta\cdot \sup\limits_{t \ge \tau \ge 0} (f(\tau)+g_{2}(t-\tau))
\end{flalign}
\end{definition}

From the above expressions, it is evident that compound tropical convolution and parallel tropical convolution can be viewed as generalized forms of tropical convolution. In other words, these operations represent advanced modes of tropical convolution, capable of degenerating to standard tropical convolution under specific conditions, i.e. when $\alpha=1$, $\beta=0$ or $\alpha=0$, $\beta=1$. 

\subsection{Specializations of compound/parallel tropical convolutions}

Different specialization schemes can be constructed by constraining the relationship between $\alpha$ and $\beta$, or by fixing their values. These specializations provide different trade-offs between representation capability, computational behavior, and parameter complexity.

Some representative specialization schemes include:

\begin{itemize}
    \item \textbf{Shared-parameter specialization} where $\beta = \alpha$ resulting in a simple weighted sum of min-plus and max-plus convolutions:
    \begin{equation}
        \alpha \cdot (f\underline{\otimes} g)
        +
        \alpha \cdot (f\overline{\otimes} g).
    \end{equation}

    \item \textbf{Opposite-sign specialization} where $\beta = -\alpha$ resulting in a contrastive combination of min-plus and max-plus convolutions:
    \begin{equation}
        \alpha \cdot (f\underline{\otimes} g)
        -
        \alpha \cdot (f\overline{\otimes} g).
    \end{equation}

    \item \textbf{Complementary-weight specialization} where $\beta = 1-\alpha$ resulting in a weighted interpolation between min-plus and max-plus convolutions:
    \begin{equation}
        \alpha \cdot (f\underline{\otimes} g)
        +
        (1-\alpha)\cdot(f\overline{\otimes} g).
    \end{equation}

    \item \textbf{Fixed-parameter specialization} where $\alpha$ and $\beta$ are set to predefined constants, such as:
    \begin{equation}
        (\alpha,\beta)\in
        \{(1,1), (0.5,0.5), (1,-1)\}.
    \end{equation}
    These configurations correspond to equal-weight combinations, normalized averaging, or contrastive combinations between min-plus and max-plus operations.
\end{itemize}

These specialization schemes can be applied to both compound tropical convolution and parallel tropical convolution. In practical implementations, the parameters may either be manually predefined or learned during training.

Definition~\ref{D:compound} and Definition~\ref{D:parallel} can be straightforwardly generalized to higher dimensions, as their mathematical expressions are general and independent of the dimensionality of the data.



\subsection{Compound tropical convolution layers}\label{subsec:ctcnn}
Based on the expressions of compound tropical convolutions, we can design and implement compound tropical convolution layers from one-dimensional to three-dimensional cases, which could improve the performance of the model in different dimensional tasks. We will introduce them in detail in the following.
Based on the setting of subsections ~\ref{subsec:tcnn} and \ref{subsec:tcnn1}, one can quickly obtain the expressions for the compound tropical convolution operation in one-dimensional, two-dimensional, and three-dimensional cases.

\begingroup
\setlength{\mathindent}{0pt}

In the one-dimensional case, the compound convolution operation can be defined as follows, namely \emph{CompoundMinPlusSumConv1d2p Layer}: \\
	\begin{flalign}
	& Y_{h,w,d,c_{\text{out}}}= \sum_{c_{\text{in}}\in[C_{\text{in}}]} \left( \alpha_{c_{\text{in}}, c_{\text{out}}} \cdot\text{MinPlus1d}(X,W, l, c_{\text{in}}, c_{\text{out}}) \right. + \nonumber \\
	& \left. \beta_{c_{\text{in}}, c_{\text{out}}} \cdot\text{MaxPlus1d}(X,W, l, c_{\text{in}}, c_{\text{out}})  \right)
	\end{flalign}

In the two-dimensional case, the compound convolution operation can be defined as follows, namely \emph{CompoundMinPlusSumConv2d2p Layer}: \\
	\begin{flalign}
		&Y_{h,w,c_{\text{out}}}= \sum_{c_{\text{in}}\in[C_{\text{in}}]} \left(\alpha_{c_{\text{in}}, c_{\text{out}}} \cdot\text{MinPlus2d}(X,W, h, w, c_{\text{in}}, c_{\text{out}}) + \right. \nonumber \\
		& \left. \beta_{c_{\text{in}}, c_{\text{out}}} \cdot \text{MaxPlus2d}(X,W, h, w, c_{\text{in}}, c_{\text{out}}) \right)
	\end{flalign}

In the three-dimensional case, the compound convolution operation can be defined as follows, namely \emph{CompoundMinPlusSumConv3d2p Layer}: \\
	\begin{flalign}
		&Y_{h,w,d,c_{\text{out}}}= \nonumber \\
		&\sum_{c_{\text{in}}\in[C_{\text{in}}]} \left(\alpha_{c_{\text{in}}, c_{\text{out}}} \cdot \text{MinPlus3d}(X,W, h, w, d, c_{\text{in}}, c_{\text{out}})+ \right. \nonumber \\
		& \left. \beta_{c_{\text{in}}, c_{\text{out}}} \cdot \text{MaxPlus3d}(X,W, h, w, d, c_{\text{in}}, c_{\text{out}}) \right)
	\end{flalign}

It's worth noting that in our compound convolutional neural network, the addition operations in MaxPlus and MinPlus are the same. That is, they share the calculation of addition, so in reality, there is only one addition for two operations.

\endgroup

\subsection{Parallel tropical convolution layers}
Similarly, based on the previous expression of parallel tropical convolution, we give the implementation of parallel tropical convolution operations. It is worth noting that parallel tropical convolution introduces more parameters and more additions on the basis of compound tropical convolution, further improving the convolution fitting ability. Compared with the standard convolutional neural network, it still reduces the amount of multiplication operations to a certain extent and remains a more efficient method for convolution calculations.

\begingroup
\setlength{\mathindent}{0pt}

In the one-dimensional case, the parallel convolution operation can be defined as follows, namely \emph{ParallelMinPlusSumConv1d2p Layer}: \\
	\begin{flalign}
	& Y_{l,c_{\text{out}}}= \sum_{c_{\text{in}}=0}^{C_{\text{in}}-1} \left(\alpha_{c_{\text{in}}, c_{\text{out}}} \times \text{MinPlus1d}(X,W_{\text{1}}, l, c_{\text{in}}, c_{\text{out}}) + \right. \nonumber \\
	& \left. \beta_{c_{\text{in}}, c_{\text{out}}} \times \text{MaxPlus1d}(X,W_{\text{2}}, l, c_{\text{in}}, c_{\text{out}}) \right)
	\end{flalign}

In the two-dimensional case, the parallel convolution operation can be defined as follows, namely \emph{ParallelMinPlusSumConv2d2p Layer}: \\
	\begin{flalign}
		&Y_{h,w,c_{\text{out}}}= \sum_{c_{\text{in}}\in[C_{\text{in}}]} \left(\alpha_{c_{\text{in}}, c_{\text{out}}} \cdot \text{MinPlus2d}(X,W_{\text{1}}, h, w, c_{\text{in}}, c_{\text{out}}) + \right. \nonumber \\
		& \left. \beta_{c_{\text{in}}, c_{\text{out}}} \cdot \text{MaxPlus2d}(X,W_{\text{2}}, h, w, c_{\text{in}}, c_{\text{out}}) \right)
	  \end{flalign}

In the three-dimensional case, the parallel convolution operation can be defined as follows, namely \emph{ParallelMinPlusSumConv3d2p Layer}: \\
	\begin{flalign}
		& Y_{h,w,d,c_{\text{out}}}= \nonumber \\
		& \sum_{c_{\text{in}}\in[C_{\text{in}}]} \left(\alpha_{c_{\text{in}}, c_{\text{out}}} \cdot \text{MinPlus3d}(X,W_{\text{1}}, h, w, d, c_{\text{in}}, c_{\text{out}}) + \right. \nonumber \\
		&  \left. \beta_{c_{\text{in}}, c_{\text{out}}} \cdot\text{MaxPlus3d}(X,W_{\text{2}}, h, w, d, c_{\text{in}}, c_{\text{out}}) \right)
		\end{flalign}

\endgroup

\section{Theoretical analysis}\label{sec:analysis}

\subsection{Complexity analysis}
According to ~\cite{howard2017mobilenets}, the standard two-dimensional convolutions have the computational cost of 
\begin{equation*}
K_{\text{h}}K_{\text{w}}C_{\text{in}}C_{\text{out}}H_{\text{out}}W_{\text{out}}
\end{equation*}
where $H_{\text{out}}$ and $W_{\text{out}}$ represent the height and width of the output feature maps, respectively. $C_{\text{in}}$ and $C_{\text{out}}$ denote the number of input and output channels, while $K_{\text{h}}$ and $K_{\text{w}}$ refer to the height and width of the convolutional kernel, respectively.
In fact, the calculation of complexity can be divided into the number of calculations for addition and the number of calculations for multiplication, then the cost of the standard two-dimensional convolution can be expressed as
\begin{align*}
	\text{multiplication: }& K_{\text{h}}K_{\text{w}}C_{\text{in}}C_{\text{out}}H_{\text{out}}W_{\text{out}} \\
	\text{addition: }& (K_{\text{h}} K_{\text{w}} C_{\text{in}} - 1) C_{\text{out}}H_{\text{out}}  W_{\text{out}} \\
	\text{comparison: }&0 
\end{align*}

In the case of tropical convolution, the cost of tropical convolution is considered in two smaller cases:

The computational cost for the MinPlusSumConv2d Layer and MaxPlusSumConv2d Layer is: 
\begin{align*}
	\text{multiplication: }&0 \\
	\text{addition: }&(C_{\text{in}} - 1 + K_{\text{h}}  K_{\text{w}}  C_{\text{in}} ) C_{\text{out}}  H_{\text{out}} W_{\text{out}}  \\
	\text{comparison: }& (K_{\text{h}}K_{\text{w}}-1)C_{\text{in}}C_{\text{out}}H_{\text{out}}W_{\text{out}} 
\end{align*}

The computational cost of other types of tropical convolutional layers is:
\begin{align*}
	\text{multiplication: }& 0 \\
   \text{addition: }& K_{\text{h}}  K_{\text{w}}  C_{\text{in}}  C_{\text{out}}  H_{\text{out}}  W_{\text{out}}  \\
   \text{comparison: }&  (K_{\text{h}}K_{\text{w}}C_{\text{in}}-1)C_{\text{out}}H_{\text{out}}W_{\text{out}} 
\end{align*}

Compared with the standard convolution, we could get the change of the number of multiplication and addition for the MinPlusSumConv2d Layer and MaxPlusSumConv2d Layer:
\begin{align*}
\text{multiplication: }&0 \\
\text{addition: }& \frac{(C_{\text{in}} - 1 + K_{\text{h}} K_{\text{w}} C_{\text{in}})  C_{\text{out}}  H_{\text{out}}  W_{\text{out}}}{(K_{\text{h}} K_{\text{w}} C_{\text{in}} - 1) C_{\text{out}}H_{\text{out}}  W_{\text{out}}}  \\ & =  1 + \frac{C_{\text{in}}}{K_{\text{h}} K_{\text{w}} C_{\text{in}} - 1}
\end{align*}
and for other types of tropical convolutional layers:
\begin{align*}
\text{multiplication: }& 0 \\
\text{addition: }&  \frac{K_{\text{h}}  K_{\text{w}}  C_{\text{in}}  C_{\text{out}}  H_{\text{out}}  W_{\text{out}} }{(K_{\text{h}} K_{\text{w}} C_{\text{in}} - 1) C_{\text{out}}H_{\text{out}}  W_{\text{out}}}  \\ & =  1 + \frac{1}{K_{\text{h}} K_{\text{w}} C_{\text{in}} - 1} \\
\end{align*}

In the case of compound tropical convolution, the cost of the compound tropical convolution operation can be expressed as
\begin{align*}
	\text{multiplication: }&   2  C_{\text{in}}  C_{\text{out}}  H_{\text{out}}  W_{\text{out}}  \\
   \text{addition: }&  ( K_{\text{h}}  K_{\text{w}}  C_{\text{in}} +  2  C_{\text{in}} -1 )  C_{\text{out}}  H_{\text{out}}  W_{\text{out}} \\
\text{comparison: }& 2(K_{\text{h}}K_{\text{w}}-1)C_{\text{in}}C_{\text{out}}H_{\text{out}}W_{\text{out}} 
\end{align*}

It is noteworthy that existing algorithms demonstrate that solving for both the maximum and minimum values in the same sequence requires only 1.5 times the cost of finding the maximum value alone. However, in our naïve implementation, we compute the maximum value twice, leading to a cost factor of 2.

Compared with the standard convolution, we could get the change of the number of multiplication and addition:
\begin{align*}
\text{multiplication: } &\frac{2  C_{\text{in}}  C_{\text{out}}  H_{\text{out}}  W_{\text{out}} }{K_{\text{h}}  K_{\text{w}}  C_{\text{in}}  C_{\text{out}}  H_{\text{out}}  W_{\text{out}}}  =  \frac{2}{K_{\text{h}}  K_{\text{w}}} \\
\text{addition: } & \frac{( K_{\text{h}}  K_{\text{w}}  C_{\text{in}} +  2  C_{\text{in}} -1 )  C_{\text{out}}  H_{\text{out}}  W_{\text{out}}}{(K_{\text{h}} K_{\text{w}} C_{\text{in}} - 1) C_{\text{out}}H_{\text{out}}  W_{\text{out}}}  \\ & = 1 + \frac{2  C_{\text{in}}}{K_{\text{h}}  K_{\text{w}}  C_{\text{in}} - 1} 
\end{align*}

In the case of parallel convolution, the cost of the parallel convolution operation can be expressed as
\begin{align*}
\text{multiplication: }&   2  C_{\text{in}}  C_{\text{out}}  H_{\text{out}}  W_{\text{out}}  \\
\text{addition: }&  ( 2  K_{\text{h}}  K_{\text{w}}  C_{\text{in}} +  2  C_{\text{in}} -1 ) C_{\text{out}}  H_{\text{out}}  W_{\text{out}}  \\
\text{comparison: }& 2(K_{\text{h}}K_{\text{w}}-1)C_{\text{in}}C_{\text{out}}H_{\text{out}}W_{\text{out}} 
\end{align*}
compared with the standard convolution, we could get the change of the number of multiplication and addition:
\begin{align*}
\text{multiplication: }&\frac{2  C_{\text{in}}  C_{\text{out}}  H_{\text{out}}  W_{\text{out}} }{K_{\text{h}} K_{\text{w}}  C_{\text{in}}  C_{\text{out}}  H_{\text{out}}  W_{\text{out}}} \\ & =  \frac{2}{K_{\text{h}}  K_{\text{w}}} \\
\text{addition: }&\frac{(2K_{\text{h}}  K_{\text{w}}  C_{\text{in}} +  2C_{\text{in}} -1)  C_{\text{out}}  H_{\text{out}}  W_{\text{out}}}{(K_{\text{h}} K_{\text{w}} C_{\text{in}} - 1) C_{\text{out}}H_{\text{out}}  W_{\text{out}}}  \\ & = 2 + \frac{2 C_{\text{in}} + 1}{K_{\text{h}} K_{\text{w}} C_{\text{in}} - 1} 
\end{align*}

\subsection{Computational analysis}
It is well-established that the computational cost of performing a multiplication on a computer is generally higher than that of an addition, while the cost of an addition is slightly higher than that of a comparison operation. However, for the sake of simplicity and fair comparison, the costs of addition and comparison can be considered approximately equivalent. Building on previous work~\citep{horowitz20141}, ~\cite{gholami2022survey} summarized the corresponding energy cost and relative area cost for different precisions under the 45~nm technology. Considering that mainstream neural network models are trained and inferred using 32-bit floating-point numbers, let us take the cost of addition and multiplication operations on 32-bit floating-point and integer numbers as an example. Their work~\citep{gholami2022survey} points out that, based on the 45~nm process, under 32-bit floating-point conditions, the relative energy cost for multiplication is 3.7~pJ, while for addition, it is 0.9~pJ. In terms of relative area cost, multiplication occupies 7700~µm², and addition occupies 4184~µm², resulting in a difference of approximately four fold. However, under 32-bit integer conditions, multiplication has an energy cost of 3.1~pJ, whereas addition costs only 0.1~pJ. The relative area cost for multiplication is 3495~µm², compared to just 137~µm² for addition, leading to an astounding 25-fold difference. That is, we can think that there is actually a proportional relationship between the cost of multiplication and addition, and we can set this scale coefficient to $\theta$, so we can give the following equation
\begin{equation}
\Omega_{u} = \theta \times \Omega_{m} + \Omega_{a} + \Omega_{c}
\label{eq:uniform}
\end{equation}
In Equation ~\ref{eq:uniform}, we define $\Omega_{\text{u}}$ as the total number of unified operations, where $\Omega_{\text{m}}$ represents the number of multiplication operations, $\Omega_{\text{a}}$ the number of addition operations, and $\Omega_{\text{c}}$ the number of comparison operations. Drawing from the data provided in~\citep{gholami2022survey}, we propose that setting the value of $\theta$ to 10 yields a metric suitable for measuring the computational cost of the model.

\section{Implementation of TCNN Framework}\label{sec:implementation}


We implemented basic TCNN, cTCNN, and pTCNN within a unified framework named \texttt{tcnn}, built on top of PyTorch~\citep{paszke2019pytorch}. To promote transparency and reproducibility, we have open-sourced our implementation on GitHub\footnote{cTCNN: \url{***}}. The details of the implementation are provided in Appendix~\ref{appendix:tcnn_framework}.

We first built the framework using a PyTorch-only pipeline. In this reference implementation, tropical convolution operators are expressed using standard PyTorch tensor primitives and trained via PyTorch automatic differentiation, which makes the framework straightforward to integrate and portable across PyTorch-supported backends (e.g., CPU and GPU). The resulting layers follow the interface conventions of PyTorch convolutions, allowing them to be used as drop-in components in common model definitions.


However, implementing tropical convolutional layers solely using the PyTorch library may result in inefficient execution due to the generation of numerous intermediate tensors and increased computational graph complexity. To address this issue, we implemented the core convolutional operations using TileLang~\citep{wang2025tilelangcomposabletiledprogramming} and integrated them into the PyTorch-based framework. This approach improves the efficiency of tropical convolutional operations while preserving the flexibility and usability of PyTorch, particularly for GPU-based training and inference.



Although the proposed GPU kernels for compound, parallel, and other tropical convolutional layers reduce memory usage and improve execution efficiency compared with naive implementations with PyTorch, current GPU architectures are primarily optimized for conventional multiply--accumulate (MAC) operations through dedicated Tensor Core hardware. In contrast, tropical convolution relies on MaxPlus and MinPlus operators, for which no dedicated hardware acceleration currently exists on modern GPUs. As a result, the computational advantages of tropical convolutional operators cannot yet be fully realized on existing hardware platforms despite the proposed kernel optimizations. Nevertheless, the proposed implementation demonstrates that tropical convolution can be effectively integrated into modern deep learning frameworks and GPU-based training pipelines. Future work may further improve the efficiency of TCNNs through hardware-aware optimization and the development of dedicated MaxPlus/MinPlus acceleration units or Tensor Core designs specialized for tropical algebra operations.


\section{Experiments}\label{sec:experiments}

In this section, we present a comparative evaluation of the proposed cTCNN and pTCNN architectures across multiple neural network models and computer vision tasks. We first introduce the evaluation metrics and datasets used throughout the experiments. Next, we investigate the effectiveness of tropical and compound tropical convolutional operators on lightweight architectures derived from LeNet. We then extend the evaluation to deeper backbone networks, including ResNet and MobileNetV2, for image classification tasks. Finally, we assess the performance of the proposed cTCNN variants on semantic segmentation tasks using U-Net-style and DeepLabV3-based architectures.

\subsection{Evaluation metric}
In order to evaluate the model we proposed, we use following metrics to analyze the performance of the method in our experiment:
\begin{itemize}
\item \textbf{accuracy}: accuracy of the model on the test set of the dataset. It is calculated as the number of correct predictions divided by the total number of predictions.
\item \textbf{AUC}: the area under the receiver operating characteristic curve (ROC AUC) of the model on the test set of the dataset. It is a measure of the performance of the model in terms of true positive rate and false positive rate.
\item \textbf{precision}: the precision of the model on the test set of the dataset. It is calculated as the number of true positive predictions divided by the total number of positive predictions.
\item  \textbf{recall}: the recall of the model on the test set of the dataset. It is calculated as the number of true positive predictions divided by the total number of actual positive instances.
\item  \textbf{F1}: the F1 score of the model on the test set of the dataset. It is the harmonic mean of precision and recall, and is a measure of the model's performance in terms of precision and recall.
\item \textbf{mIoU}: the mean Intersection over Union (mIoU) of the model on the test set of the dataset. It is a measure of the performance of the model in terms of the overlap between the predicted and actual segmentation masks.
\end{itemize}

\subsection{Dataset}
\begin{figure}
	\centering
	\includegraphics[scale=0.18]{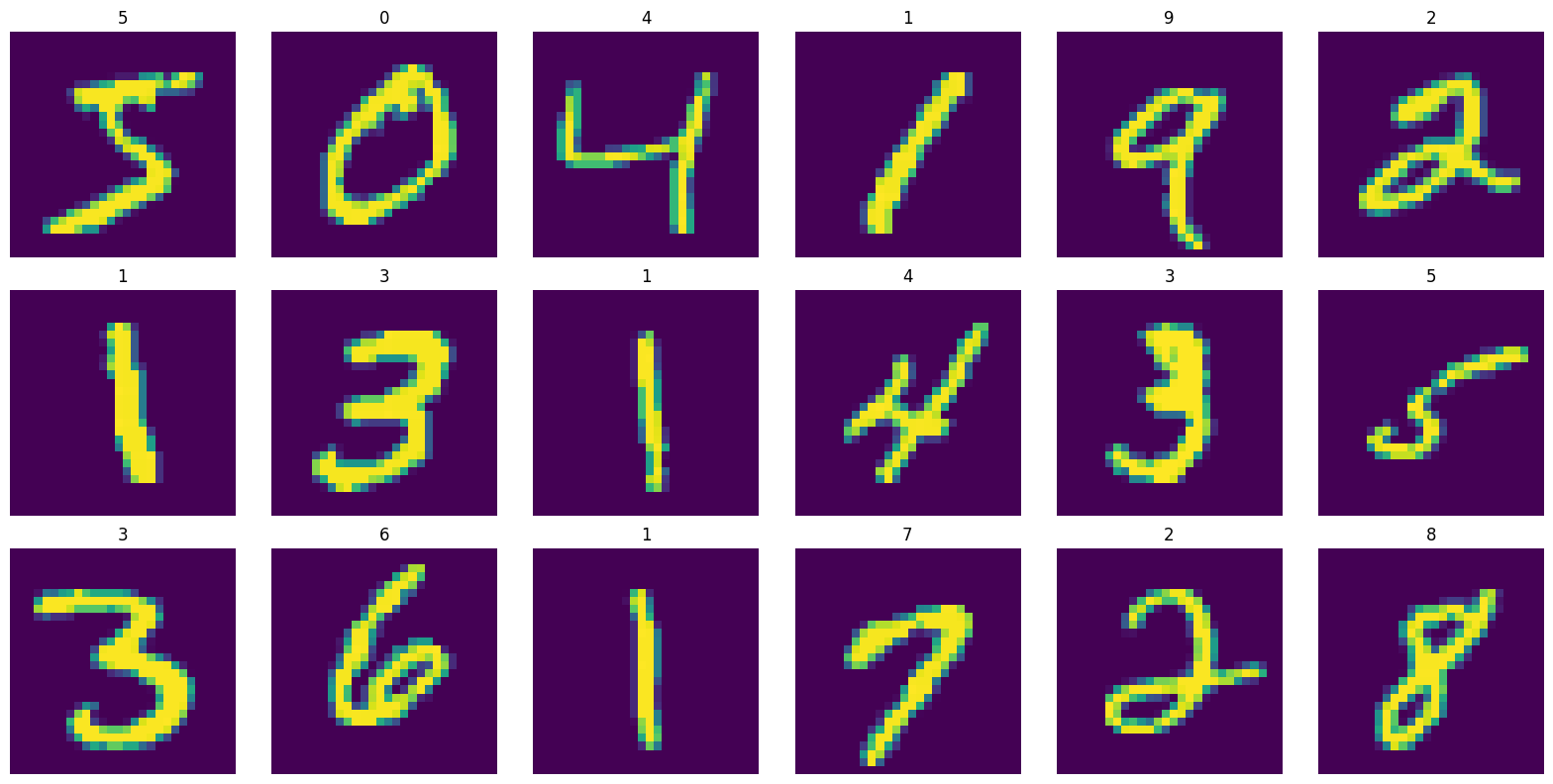}\\
	[10pt]
	\includegraphics[scale=0.18]{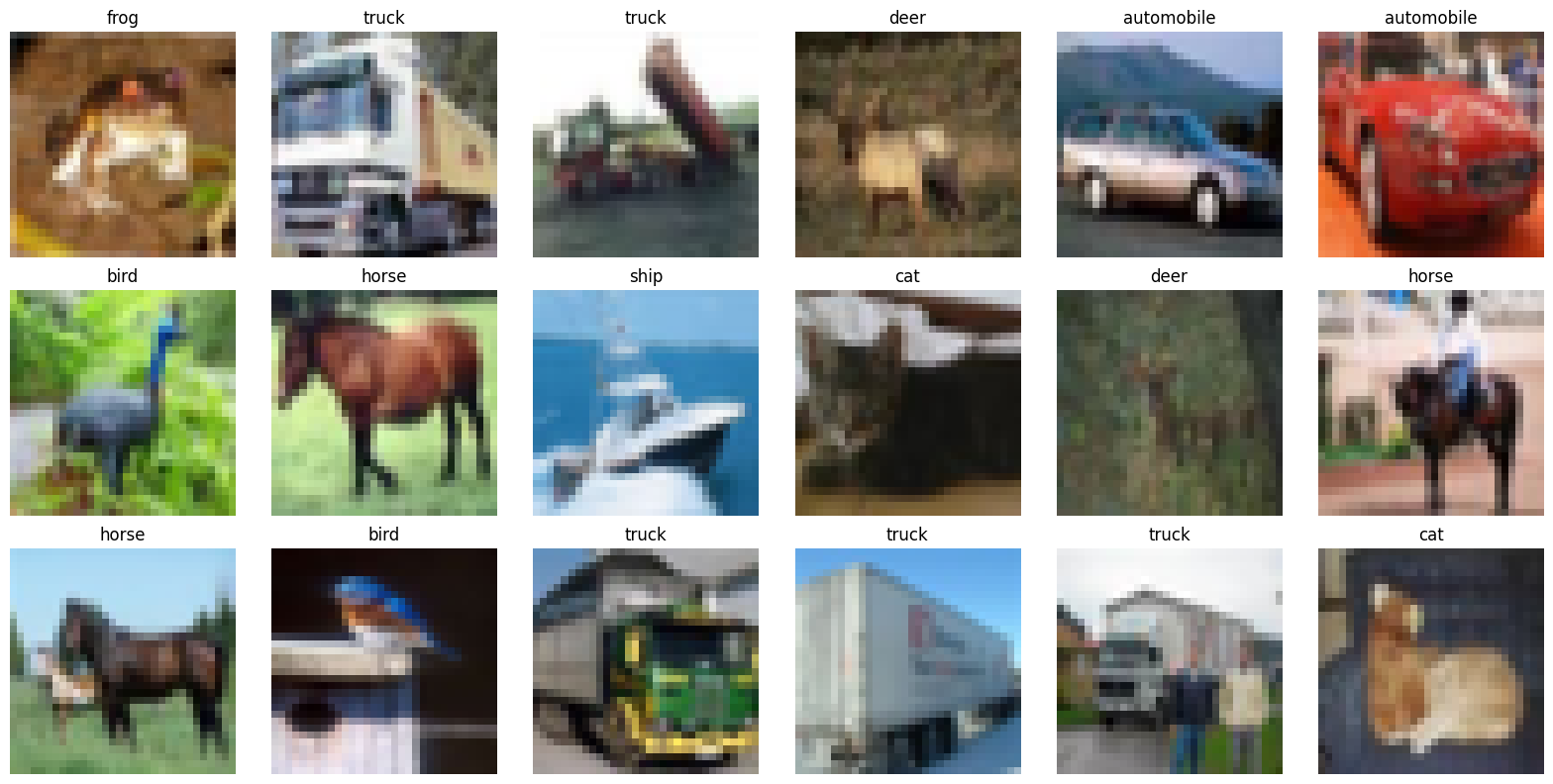}
\caption{Samples of the MNIST dataset and the CIFAR-10 dataset, showing the diversity of images in each dataset.  (top) MNIST dataset samples, showcasing handwritten digits from 0 to 9. (bottom) CIFAR-10 dataset samples, illustrating various objects including animals and vehicles.}
\label{images:image_sample}
\end{figure}

The MNIST~\citep{lecun1998gradient} dataset is a widely used benchmark dataset for image classification tasks. It consists of 60,000 grayscale images of handwritten digits from 0 to 9, with each image being a 28x28 pixel square. The dataset is commonly used to train and evaluate machine learning models for digit recognition. Some samples of the MNIST dataset are shown in Figure ~\ref{images:image_sample}.

The FashionMNIST~\citep{xiao2017fashion} dataset is a widely used benchmark dataset for image classification tasks. It consists of 60,000 grayscale images of fashion items, with 10 different classes including T-shirts/tops, trousers, pullovers, dresses, coats, sandals, shirts, sneakers, bags, and ankle boots. Each image is a 28x28 pixel square. The dataset is commonly used to evaluate the performance of image classification models on fashion items.

The SVHN (Street View House Numbers)~\citep{netzer2011reading} dataset is a large-scale dataset for digit recognition in natural images. It consists of over 600,000 color images of house numbers collected from Google Street View. The dataset is more challenging than MNIST as it contains more variations in digit appearance, such as different fonts, sizes, and orientations. It is commonly used to evaluate the robustness of digit recognition models.

The CIFAR-10/100~\citep{krizhevsky2009learning} datasets are other popular benchmark datasets for image classification. The CIFAR-10 dataset consists of 60,000 color images, with 10 different classes including airplanes, automobiles, birds, cats, deer, dogs, frogs, horses, ships, and trucks. Each image is a 32x32 pixel square. The dataset is commonly used to evaluate the performance of image classification models. While the CIFAR-100 dataset is similar to CIFAR-10, it contains 100 classes with 600 images each, making it a more challenging benchmark for image classification tasks. Some samples of the CIFAR-10 dataset are shown in Figure ~\ref{images:image_sample}. Some samples of the MNIST dataset and the CIFAR-10 dataset are shown in Figure ~\ref{images:image_sample}.

Tiny ImageNet is a subset of the ImageNet dataset~\citep{deng2009imagenet, russakovsky2015imagenet} introduced for efficient benchmarking of image classification models under limited computational resources. It contains 200 object categories, with each category consisting of 500 training images, 50 validation images, and 50 test images. All images are resized to a resolution of $64\times64$ pixels, making the dataset significantly smaller and computationally lighter than the original ImageNet while still preserving substantial visual diversity and classification complexity. Tiny ImageNet is widely used for evaluating lightweight neural networks, training strategies, and novel deep learning architectures in computer vision research.

The CamVid~\citep{BROSTOW200988camvid} dataset is a road scene semantic segmentation benchmark containing video sequences captured from the perspective of a driving vehicle. The dataset consists of 701 pixel-wise annotated images with a resolution of $720 \times 960$ and originally defines 11 semantic categories (e.g., sky, road, building, etc.). The CamVid dataset has 367 images used for training, 101 for validation, and 233 for testing.



The PASCAL VOC dataset is a widely adopted benchmark for semantic segmentation and object recognition tasks~\citep{everingham2010pascal,pascalvoc2012}. In this work, the training and validation sets were constructed by combining images from the PASCAL VOC 2007 and VOC 2012 segmentation datasets, resulting in 1,673 training images and 1,662 validation images. The VOC 2007 segmentation test set containing 210 images was used for evaluation. The dataset defines 20 foreground object categories together with one background class.


\subsection{Classification Task with LeNet}\label{sec:2dc-TCNN}

\subsubsection{Experimental Setup}
In this subsection, we would focus on two-dimensional image classification, while 1D and 3D tasks are discussed in Appendix~\ref{sec:1dc-TCNN} and~\ref{sec:3dc-TCNN}. The models used for the two-dimensional image classification task is shown as table ~\ref{tab:overall-architecture} and they just were set as the configuration as 2d in table and were slightly tuned to fit different datasets. We use Adam whose initial learning rate is 0.001 as the optimizer of all models and we also use exponentialLR scheduler with a decay factor (gamma) of 0.9 for all models. This scheduler gradually decreases the learning rate during training to improve convergence and performance. We set batch size as 64. Similarly, because too large epochs can lead to overfitting, based on our convergence analysis of a single comparison experiment, we set the epochs to 100 for the MNIST dataset and 50 for the other datasets. We still repeated the experiment 5 times for each model and reported the mean and variance of the experimental metrics.

\begin{table*}[t]
	\centering
	\caption{The architecture of the LeNet models defined in our experiment}
	\label{tab:overall-architecture}
	\begingroup
	\setlength{\tabcolsep}{3pt}
	\renewcommand{\arraystretch}{0.9}
	\scriptsize
	\resizebox{\textwidth}{!}{%
	\begin{tabular}{|l|l|l|l|l|l|l|l|l|l|l|}
	\hline
	\diagbox{Models}{Layers}  & \begin{tabular}[l]{@{}l@{}}1d:{[}80/6,2{]}\\ 2d:{[}5/6,2{]}\\ 3d:{[}5/6,2{]}\end{tabular} & function & \begin{tabular}[l]{@{}l@{}}1d:{[}2,2{]}\\ 2d:{[}2,2{]}\\ 3d:{[}2,2{]}\end{tabular}  & \begin{tabular}[l]{@{}l@{}}1d:{[}3/16{]}\\ 2d:{[}5/16{]}\\ 3d:{[}5/16{]}\end{tabular} & function &  \begin{tabular}[l]{@{}l@{}}1d:{[}2,2{]}\\ 2d:{[}2,2{]}\\ 3d:{[}2,2{]}\end{tabular} & &(n,20)&(20,84)&(84,c)\\ \hline

	 LeNet& Conv & Sigmoid & \multirow{13}{*}{AP}  &  Conv  & Sigmoid  & \multirow{13}{*}{AP}  &  \multirow{12}{*}{Flatten} & L(S) & L(S) & L(S)\\ \cline{1-3} \cline{5-6} \cline{9-11} 
	 LeNet-ReLU & Conv & ReLU & & Conv & ReLU  &  &  & \multirow{12}{*}{L(R)} & \multirow{12}{*}{L(R)} & \multirow{12}{*}{L(R)}  \\ \cline{1-3} \cline{5-6} 
	 LeNet-F-\Rmnum{1}& MinPusSumConv & \multirow{11}{*}{}  &  & MaxPlusSumConv & \multirow{12}{*}{} &    &  &  &  &  \\ \cline{1-2} \cline{5-5}
	 LeNet-F-\Rmnum{2}& MinPlusMaxConv &  &  & MaxPlusMinConv &  &  &  &  &  &  \\ \cline{1-2} \cline{5-5} 
	 LeNet-F/M-\Rmnum{3}& MinPluSumConv &  &  & Conv &  &  &  &  &  &  \\ \cline{1-2} \cline{5-5}
	 LeNet-C-$\alpha$& CompoundConv1p &  &  &CompoundConv1p  &  &  &  &  &  &  \\ \cline{1-2} \cline{5-5}
	 LeNet-C-$\alpha / \beta$ & CompoundConv2p &  &  &CompoundConv2p  &  &  &  &  &  &  \\ \cline{1-2} \cline{5-5}
	 LeNet-CM-$\alpha$ & CompoundConv1p &  &  &Conv  &  &  &  &  &  &  \\ \cline{1-2} \cline{5-5}
	 LeNet-CM-$\alpha / \beta$ & CompoundConv2p &  &  &Conv  &  &  &  &  &  &  \\ \cline{1-2} \cline{5-5}
	 LeNet-P-$\alpha$ & ParallelConv1p &  &  & ParallelConv1p  &  &  &  &  &  &  \\ \cline{1-2} \cline{5-5}
	 LeNet-P-$\alpha / \beta$ & ParallelConv2p &  &  & ParallelConv2p  &  &  &  &  &  &  \\ \cline{1-2} \cline{5-5}
	 LeNet-PM-$\alpha$  & ParallelConv1p &  &  & Conv &  &  &  &  &  &  \\ \cline{1-2} \cline{5-5}
	 LeNet-PM-$\alpha / \beta$ & ParallelConv2p &  &  & Conv &  &  &  &  &  &  \\ \cline{1-2} \cline{5-5}
	 \hline
	\end{tabular}
	}%
	\endgroup
	\par\smallskip
	\begin{minipage}{\textwidth}
		\raggedright\footnotesize
		1d:[80/6,2] means the layer is a one-dimensional convolutional layer, the kernel size is 80, the number of output channels is 6, the stride is 2. 2d:[5/6,2] and 3d:[5/6,2] are similar to 1d. 
		AP is the abbreviation of the parallel tropical convolutional layer. 1d:[2,2] means the layer is a average pooling layer, the kernel size is 2, the stride is 2. 2d:[2,2] and 3d:[2,2] are similar to 1d.
		L(S) is the abbreviation of the linear layer with the sigmoid activation function. L(R) is the abbreviation of the linear layer with the ReLU activation function.  CompoundConv1p and ParallelConv1p are abbreviations for the CompoundMinMaxPlusSumConv1p and ParallelMinMaxPlusSumConv1p layers, respectively, and both correspond to the complementary-weight specialization ($\beta = 1-\alpha$). CompoundConv2p and ParallelConv2p are the abbreviations for the CompoundMinMaxPlusSumConv2p and ParallelMinMaxPlusSumConv2p layers, respectively. n is the dimension of the output tensor of the Flatten layer. c is the number of classes of the classification task. LeNet-F stands for the LeNet model derived from tropical convolution proposed by Fan et al~\cite{fan2021alternative}, LeNet-C represents the LeNet-Compound model, and LeNet-P represents the LeNet-Parallel model. M represents a hybrid structure of tropical convolutional and standard convolution. $\alpha$ and $\beta$ denote the parametric forms of the included compound or parallel tropical convolution, respectively.
	\end{minipage}

\end{table*}


\begin{table*}[htbp]
	\centering
	\caption{Performance comparison of various LeNet-based models across MNIST, Fashion MNIST, CIFAR-10, and SVHN datasets. The table reports accuracy, AUC (Area Under the Curve), F1 score, precision, recall, number of operations (Ops$_{u}$ with $\theta = 10$), and the number of parameters for each model.}
	\begingroup
	\setlength{\tabcolsep}{3pt}
	\renewcommand{\arraystretch}{0.95}
	\scriptsize
	\resizebox{\textwidth}{!}{%
	\begin{tabular}{llcccccccc}
	\toprule
	Dataset & Model & Accuracy &Auc&F1&Precision&Recall& $\Omega_{u}$($\theta$ = 10) & Parameters \\
	\midrule
MNIST &LeNet-P-$\alpha$ & 98.95 $ \pm $ 0.19 & 1.00 $ \pm $ 0.00 & 0.99 $ \pm $ 0.00 & 0.99 $ \pm $ 0.00 & 0.99 $ \pm $ 0.00 & \num{1.89e7} & 64358\\
	&LeNet-P-$\alpha / \beta$  & 98.96 $ \pm $ 0.05 & 1.00 $ \pm $ 0.00 & 0.99 $ \pm $ 0.00 & 0.99 $ \pm $ 0.00 & 0.99 $ \pm $ 0.00 & \num{1.89e7}  & 64460\\
	&LeNet-PM-$\alpha$ & 98.68 $ \pm $ 0.12 & 1.00 $ \pm $ 0.00 & 0.99 $ \pm $ 0.00 & 0.99 $ \pm $ 0.00 & 0.99 $ \pm $ 0.00 &\num{4.57e7}  & 61862\\
	&LeNet-PM-$\alpha / \beta$ & 98.71 $ \pm $ 0.13 & 1.00 $ \pm $ 0.00 & 0.99 $ \pm $ 0.00 & 0.99 $ \pm $ 0.00 & 0.99 $ \pm $ 0.00 & \num{4.57e7} & 61868\\
	&LeNet-C-$\alpha$ & 98.86 $ \pm $ 0.10 & 1.00 $ \pm $ 0.00 & 0.99 $ \pm $ 0.00 & 0.99 $ \pm $ 0.00 & 0.99 $ \pm $ 0.00 & \num{1.43e7} & 61808\\
	&LeNet-C-$\alpha / \beta$ & 98.95 $ \pm $ 0.07 & 1.00 $ \pm $ 0.00 & 0.99 $ \pm $ 0.00 & 0.99 $ \pm $ 0.00 & 0.99 $ \pm $ 0.00 & \num{1.43e7} & 61910\\
	&LeNet-CM-$\alpha$ & 98.46 $ \pm $ 0.23 & 1.00 $ \pm $ 0.00 & 0.98 $ \pm $ 0.00 & 0.98 $ \pm $ 0.00 & 0.98 $ \pm $ 0.00 & \num{4.50e7} & 61712\\
	&LeNet-CM-$\alpha / \beta$ & 98.53 $ \pm $ 0.11 & 1.00 $ \pm $ 0.00 & 0.99 $ \pm $ 0.00 & 0.99 $ \pm $ 0.00 & 0.99 $ \pm $ 0.00 & \num{4.50e7} & 61718\\
	&LeNet-F-\Rmnum{1} & 98.86 $ \pm $ 0.05 & 1.00 $ \pm $ 0.00 & 0.99 $ \pm $ 0.00 & 0.99 $ \pm $ 0.00 & 0.99 $ \pm $ 0.00 & \num{0.98e7} & 61706\\
	&LeNet-F-\Rmnum{2} & 98.96 $ \pm $ 0.06 & 1.00 $ \pm $ 0.00 & 0.99 $ \pm $ 0.00 & 0.99 $ \pm $ 0.00 & 0.99 $ \pm $ 0.00 & \num{0.98e7} & 61706\\
	&LeNet-F-\Rmnum{3} & 98.42 $ \pm $ 0.17 & 1.00 $ \pm $ 0.00 & 0.98 $ \pm $ 0.00 & 0.98 $ \pm $ 0.00 & 0.98 $ \pm $ 0.00 & \num{4.43e7} & 61706 \\
	&LeNet & 98.59 $ \pm $ 0.09 & 1.00 $ \pm $ 0.00 & 0.99 $ \pm $ 0.00 & 0.99 $ \pm $ 0.00 & 0.99 $ \pm $ 0.00  &\num{5.07e7} & 61706\\
	&LeNet-ReLU & \textbf{98.99 $ \pm $ 0.07} & 1.00 $ \pm $ 0.00 & 0.99 $ \pm $ 0.00 & 0.99 $ \pm $ 0.00 & 0.99 $ \pm $ 0.00 & \num{5.07e7} & 61706 \\          
	\midrule
	\multirow{2}{*}{\begin{tabular}{@{}c@{}}Fashion \\ MNIST\end{tabular}}
	 &LeNet-P-$\alpha$ & 90.61 $ \pm $ 0.09 & 0.99 $ \pm $ 0.00 & 0.91 $ \pm $ 0.00 & 0.91 $ \pm $ 0.00 & 0.91 $ \pm $ 0.00 & \num{1.89e7} & 64358\\
&LeNet-P-$\alpha / \beta$ & \textbf{91.11 $ \pm $ 0.18} & 0.99 $ \pm $ 0.00 & 0.91 $ \pm $ 0.00 & 0.91 $ \pm $ 0.00 & 0.91 $ \pm $ 0.00 & \num{1.89e7} & 64460\\
&LeNet-PM-$\alpha$ & 90.57 $ \pm $ 0.21 & 0.99 $ \pm $ 0.00 & 0.91 $ \pm $ 0.00 & 0.91 $ \pm $ 0.00 & 0.91 $ \pm $ 0.00 & \num{4.57e7} & 61862\\
&LeNet-PM-$\alpha / \beta$ & 90.53 $ \pm $ 0.21 & 0.99 $ \pm $ 0.00 & 0.91 $ \pm $ 0.00 & 0.91 $ \pm $ 0.00 & 0.91 $ \pm $ 0.00 & \num{4.57e7} & 61868\\
&LeNet-C-$\alpha$ & 90.41 $ \pm $ 0.34 & 0.99 $ \pm $ 0.00 & 0.90 $ \pm $ 0.00 & 0.90 $ \pm $ 0.00 & 0.90 $ \pm $ & \num{1.43e7} & 61808\\
&LeNet-C-$\alpha / \beta$ & 90.83 $ \pm $ 0.19 & 0.99 $ \pm $ 0.00 & 0.91 $ \pm $ 0.00 & 0.91 $ \pm $ 0.00 & 0.91 $ \pm $ 0.00 & \num{1.43e7} & 61910\\
&LeNet-CM-$\alpha$ & 90.59 $ \pm $ 0.19 & 0.99 $ \pm $ 0.00 & 0.91 $ \pm $ 0.00 & 0.91 $ \pm $ 0.00 & 0.91 $ \pm $ 0.00 & \num{4.50e7} & 61712\\
&LeNet-CM-$\alpha / \beta$ & 90.18 $ \pm $ 0.15 & 0.99 $ \pm $ 0.00 & 0.90 $ \pm $ 0.00 & 0.90 $ \pm $ 0.00 & 0.90 $ \pm $ 0.00 & \num{4.50e7} & 61718\\
&LeNet-F-\Rmnum{1} & 88.13 $ \pm $ 0.17 & 0.99 $ \pm $ 0.00 & 0.88 $ \pm $ 0.00 & 0.88 $ \pm $ 0.00 & 0.88 $ \pm $ & \num{0.98e7} & 61706\\
&LeNet-F-\Rmnum{2} & 89.16 $ \pm $ 0.23 & 0.99 $ \pm $ 0.00 & 0.89 $ \pm $ 0.00 & 0.89 $ \pm $ 0.00 & 0.89 $ \pm $ 0.00 & \num{0.98e7} & 61706\\
&LeNet-F-\Rmnum{3} & 90.23 $ \pm $ 0.13 & 0.99 $ \pm $ 0.00 & 0.90 $ \pm $ 0.00 & 0.90 $ \pm $ 0.00 & 0.90 $ \pm $ 0.00 & \num{4.43e7} & 61706 \\
&LeNet & 86.03 $ \pm $ 0.07 & 0.99 $ \pm $ 0.00 & 0.86 $ \pm $ 0.00 & 0.86 $ \pm $ 0.00 & 0.86 $ \pm $ 0.00 &\num{5.07e7}& 61706\\
&LeNet-ReLU & 90.12 $ \pm $ 0.27 & 0.99 $ \pm $ 0.00 & 0.90 $ \pm $ 0.00 & 0.90 $ \pm $ 0.00 & 0.90 $ \pm $ 0.00 & \num{5.07e7} & 61706 \\
	\midrule
	SVHN &LeNet-P-$\alpha$ & 88.61 $ \pm $ 0.36 & 0.99 $ \pm $ 0.00 & 0.87 $ \pm $ 0.00 & 0.87 $ \pm $ 0.00 & 0.87 $ \pm $ 0.00 & \num{3.41e7} & 86090\\
&LeNet-P-$\alpha / \beta$ & \textbf{88.86 $ \pm $ 0.21} & 0.99 $ \pm $ 0.00 & 0.88 $ \pm $ 0.00 & 0.88 $ \pm $ 0.00 & 0.88 $ \pm $ 0.00  & \num{3.41e7} & 86204\\
&LeNet-PM-$\alpha$ & 87.11 $ \pm $ 0.60 & 0.99 $ \pm $ 0.00 & 0.86 $ \pm $ 0.01 & 0.86 $ \pm $ 0.01 & 0.86 $ \pm $ 0.01 & \num{7.28e7} & 83594\\
&LeNet-PM-$\alpha / \beta$ & 88.63 $ \pm $ 0.31 & 0.99 $ \pm $ 0.00 & 0.88 $ \pm $ 0.00 & 0.87 $ \pm $ 0.00 & 0.88 $ \pm $ 0.00 & \num{7.28e7} & 83612\\
&LeNet-C-$\alpha$ & 88.41 $ \pm $ 0.29 & 0.99 $ \pm $ 0.00 & 0.87 $ \pm $ 0.00 & 0.87 $ \pm $ 0.00 & 0.87 $ \pm $ 0.00 & \num{2.58e7} & 83240\\
&LeNet-C-$\alpha / \beta$ & 88.54 $ \pm $ 0.28 & 0.99 $ \pm $ 0.00 & 0.87 $ \pm $ 0.00 & 0.87 $ \pm $ 0.00 & 0.87 $ \pm $ 0.00 & \num{2.58e7} & 83354\\
&LeNet-CM-$\alpha$ & 87.27 $ \pm $ 0.77 & 0.99 $ \pm $ 0.00 & 0.86 $ \pm $ 0.01 & 0.86 $ \pm $ 0.01 & 0.86 $ \pm $ 0.01 & \num{7.00e7} & 83144\\
&LeNet-CM-$\alpha / \beta$ & 88.48 $ \pm $ 0.30 & 0.99 $ \pm $ 0.00 & 0.87 $ \pm $ 0.00 & 0.87 $ \pm $ 0.00 & 0.88 $ \pm $ 0.00 & \num{7.00e7} & 83162\\
&LeNet-F-\Rmnum{1} & 84.38 $ \pm $ 0.40 & 0.98 $ \pm $ 0.00 & 0.83 $ \pm $ 0.00 & 0.83 $ \pm $ 0.00 & 0.83 $ \pm $ 0.00 & \num{1.75e7} & 83126\\
&LeNet-F-\Rmnum{2} & 85.72 $ \pm $ 0.26 & 0.98 $ \pm $ 0.00 & 0.84 $ \pm $ 0.00 & 0.85 $ \pm $ 0.00 & 0.84 $ \pm $ 0.00 & \num{1.75e7} & 83126\\
&LeNet-F-\Rmnum{3} & 85.32 $ \pm $ 0.62 & 0.98 $ \pm $ 0.00 & 0.84 $ \pm $ 0.01 & 0.84 $ \pm $ 0.01 & 0.84 $ \pm $ 0.01& \num{6.73e7} & 83126 \\
&LeNet & 40.67 $ \pm $ 25.98 & 0.70 $ \pm $ 0.20 & 0.30 $ \pm $ 0.31 & 0.30 $ \pm $ 0.33 & 0.33 $ \pm $ 0.28 & \num{9.22e7} & 83126\\
&LeNet-ReLU & 88.76 $ \pm $ 0.31 & 0.99 $ \pm $ 0.00 & 0.88 $ \pm $ 0.00 & 0.88 $ \pm $ 0.00 & 0.88 $ \pm $ 0.00 & \num{9.22e7} & 83126\\
	\midrule
	CIFAR-10  &LeNet-P-$\alpha$ & 61.16 $ \pm $ 0.66 & 0.93 $ \pm $ 0.00 & 0.61 $ \pm $ 0.01 & 0.61 $ \pm $ 0.01 & 0.61 $ \pm $ 0.01 &\num{3.41e7} & 86090\\
&LeNet-P-$\alpha / \beta$ & \textbf{66.82 $ \pm $ 0.52} & 0.94 $ \pm $ 0.00 & 0.67 $ \pm $ 0.01 & 0.67 $ \pm $ 0.01 & 0.67 $ \pm $ 0.01 &\num{3.41e7} & 86204\\
&LeNet-PM-$\alpha$  & 64.95 $ \pm $ 0.92 & 0.94 $ \pm $ 0.00 & 0.65 $ \pm $ 0.01 & 0.65 $ \pm $ 0.01 & 0.65 $ \pm $ 0.01 & \num{7.28e7} & 83594\\
&LeNet-PM-$\alpha / \beta$ & 64.49 $ \pm $ 0.82 & 0.94 $ \pm $ 0.00 & 0.64 $ \pm $ 0.01 & 0.64 $ \pm $ 0.01 & 0.64 $ \pm $ 0.01 &\num{7.28e7}  & 83612\\
&LeNet-C-$\alpha$ & 60.65 $ \pm $ 0.59 & 0.93 $ \pm $ 0.00 & 0.60 $ \pm $ 0.01 & 0.60 $ \pm $ 0.01 & 0.61 $ \pm $ 0.01 &\num{2.58e7}  & 83240\\
&LeNet-C-$\alpha / \beta$ & 65.90 $ \pm $ 0.65 & 0.94 $ \pm $ 0.00 & 0.66 $ \pm $ 0.01 & 0.66 $ \pm $ 0.01 & 0.66 $ \pm $ 0.01 &\num{2.58e7}  & 83354\\
&LeNet-CM-$\alpha$ & 63.34 $ \pm $ 1.30 & 0.93 $ \pm $ 0.00 & 0.63 $ \pm $ 0.01 & 0.63 $ \pm $ 0.01 & 0.63 $ \pm $ 0.01 &\num{7.00e7}  & 83144\\
&LeNet-CM-$\alpha / \beta$ & 64.59 $ \pm $ 0.57 & 0.94 $ \pm $ 0.00 & 0.64 $ \pm $ 0.01 & 0.64 $ \pm $ 0.01 & 0.65 $ \pm $ 0.01 &\num{7.00e7}  & 83162\\
&LeNet-F-\Rmnum{1} & 50.73 $ \pm $ 0.49 & 0.88 $ \pm $ 0.00 & 0.50 $ \pm $ 0.00 & 0.50 $ \pm $ 0.01 & 0.51 $ \pm $ 0.00 &\num{1.75e7}  & 83126\\
&LeNet-F-\Rmnum{2} & 60.50 $ \pm $ 0.80 & 0.92 $ \pm $ 0.00 & 0.60 $ \pm $ 0.01 & 0.60 $ \pm $ 0.01 & 0.60 $ \pm $ 0.01 &\num{1.75e7}  & 83126\\
&LeNet-F-\Rmnum{3} & 62.27 $ \pm $ 0.73 & 0.93 $ \pm $ 0.00 & 0.62 $ \pm $ 0.01 & 0.62 $ \pm $ 0.01 & 0.62 $ \pm $ 0.01 &\num{6.73e7}  & 83126\\
&LeNet & 48.00 $ \pm $ 2.53 & 0.88 $ \pm $ 0.01 & 0.47 $ \pm $ 0.03 & 0.47 $ \pm $ 0.03 & 0.48 $ \pm $ 0.03 &\num{9.22e7} & 83126\\
&LeNet-ReLU & 62.49 $ \pm $ 0.67 & 0.93 $ \pm $ 0.00 & 0.62 $ \pm $ 0.01 & 0.62 $ \pm $ 0.01 & 0.62 $ \pm $ 0.01 &\num{9.22e7} & 83126\\
	\bottomrule
	\end{tabular}%
	}
	\endgroup
	\label{tab:2dc-TCNN-results}
\end{table*}

\subsubsection{Results}
Table ~\ref{tab:2dc-TCNN-results} shows the experimental results on the MNIST, FashionMNIST, CIFAR-10, and SVHN datasets. The metrics evaluated include Accuracy, AUC, F1 score, Precision, Recall, $\Omega_{u}$, and Parameters. 

In the MNIST dataset, almost all models achieve near-perfect performance, with accuracy, AUC, F1 score, precision, and recall values close to 1.00. The LeNet-ReLU model achieves a perfect score across all metrics, indicating its capability to effectively handle this simple task. This is not surprising, as LeNet was originally designed for the MNIST dataset. However, simpler models like LeNet-P-$\alpha$ and LeNet-P-$\alpha/\beta$, while also achieving high performance, demonstrate slightly higher computational costs (1.89 $\times 10^7$ $\Omega_{u}$), compared to more efficient models like LeNet-C-$\alpha$ (1.43 $\times 10^7$  $\Omega_{u}$), which balances accuracy and computational efficiency.

In the Fashion MNIST dataset, the LeNet-P-$\alpha / \beta$ variants again perform well, achieving up to 91.11\%. This shows they can handle slightly harder data. The computational cost across these models is relatively similar to those seen in MNIST, but the increased dataset complexity does result in a small drop in accuracy for some models, such as LeNet-ReLU, which scores an accuracy of 90.12\%, slightly lower than in the MNIST dataset. Despite the drop, the models based on tropical convolution remain computationally efficient, showing a good balance of accuracy and computational demands, which still matches or even exceeds the performance of LeNet-ReLU.


On the SVHN dataset, the best-performing models, LeNet-P-$\alpha / \beta$ and LeNet-PM-$\alpha / \beta$, achieve accuracies of 88.86\% and 88.63\%, respectively. These models also maintain computational efficiency, with operations ranging from 3.41 $\times 10^7$ to 7.28 $\times 10^7$ $\Omega_{u}$. However, the complexity of the dataset leads to notable reductions in the F1 scores, precision, and recall across all models. Even though overall performance decreases, several tropical variants remain competitive and achieve the top accuracy on this dataset.

On the more challenging CIFAR-10 dataset, performance discrepancies between models become more pronounced. Here, the accuracy of models drops further, with the highest being 66.82\% for LeNet-P-$\alpha / \beta$. The computational cost remains consistent with what is seen in SVHN, but the performance drop across all metrics, including AUC, F1 score, and recall, indicates the difficulty the models face in extracting meaningful features from the more complex CIFAR-10  dataset. The LeNet-ReLU model, in particular, shows a steep decline in performance, achieving an accuracy of only 62.49\% with a high computational cost of 9.22 $\times 10^7$ Ops$_{u}$, demonstrating that this model struggles with the increased dataset complexity.

Overall, the results demonstrate the adaptability of LeNet-based models to varying data complexity. While simpler datasets like MNIST allow almost all models to achieve near-perfect performance, more complex datasets like CIFAR-10 and SVHN reveal the limitations of certain models, particularly as the dataset complexity increases. The LeNet-P-$\alpha / \beta$ models emerge as the most balanced, offering high accuracy with moderate computational costs across all datasets. Meanwhile, models like LeNet-ReLU, while performing well on simpler tasks, may require further optimization or architectural adjustments to handle more challenging datasets effectively. These experiments suggest that tropical-convolution-based models can provide a favorable robustness and accuracy--efficiency trade-off across multiple image-classification benchmarks.

\subsection{Exploring Deeper Networks}\label{subsec:deepnet}

In the previous subsection, we reported experimental results for LeNet-style networks using tropical convolution operators. Since CNNs have evolved substantially beyond LeNet and remain fundamental to modern architectures such as ResNet~\citep{he_deep_2016} and MobileNetV2~\citep{sandler2018mobilenetv2}, we next study how tropical operators behave when integrated into deeper backbones.

For earlier deep models such as AlexNet~\citep{krizhevsky2012imagenet} and VGG~\citep{simonyan2014very}, we do not attempt an exhaustive optimization study. Instead, we focus on mainstream architectures that are built by systematically stacking reusable blocks, including the ResNet, DenseNet, and MobileNet families (the latter using an efficient inverted residual structure). Considering dataset size and computational constraints, we select ResNet18, ResNet34, and MobileNetV2 as benchmarks. In this section, we employ both $\alpha$ and $\beta$ as the learnable parameters for the minplus and maxplus operations within the tropical convolutional layers; a detailed ablation study examining the effect of this design choice on MobileNetV2 is provided in Appendix~\ref{subsec:ablationStudyParametersOnMobilenet}.

\subsubsection{ResNet Models}

\begin{figure}
\centering
\includegraphics[width=\linewidth]{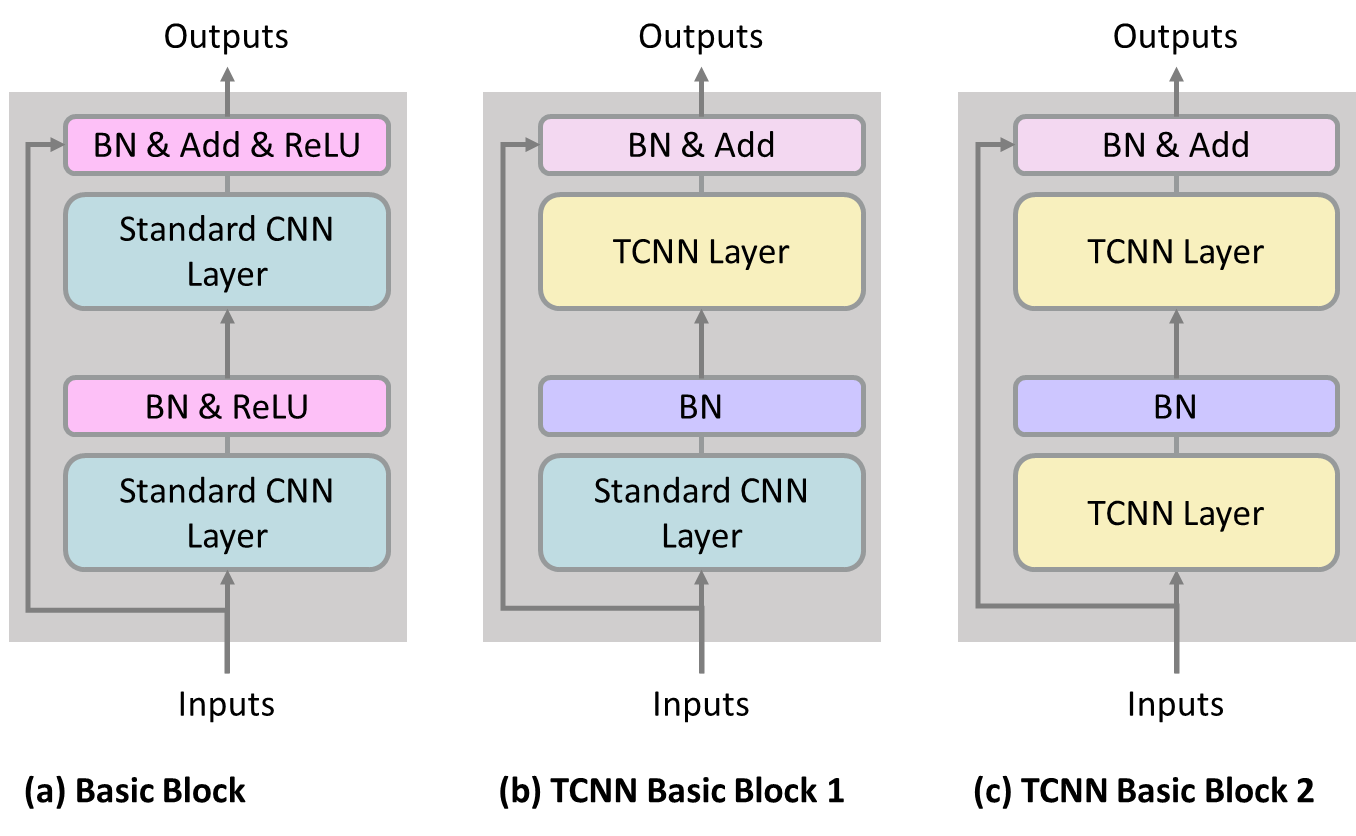}
\caption{Diagram of Different Basic Blocks
(a) Basic block: Contains a convolutional layer (Conv), batch normalization layer (BN), ReLU activation function, another convolutional layer, and batch normalization layer. (b) TCNN basic block 1: Contains a convolutional layer, batch normalization layer, compound / parallel convolutional layer (CompoundConv / ParallelConv), and batch normalization layer. (c) TCNN basic block 2: Contains two compound / parallel convolutional layers and batch normalization layers.}
\label{images:basic_block_overview}
\end{figure}

As noted earlier, ResNet is built upon a block-stacked network structure, with ResNet18 and ResNet34 primarily composed of BasicBlocks. To integrate tropical algebra into ResNet, we propose new BasicBlocks based on tropical algebra. Specifically, we introduce TCNN Basic Block 1 and TCNN Basic Block 2 as illustrated in Figure ~\ref{images:basic_block_overview}.


We construct two TCNN-style BasicBlock variants via layer substitution (Figure~\ref{images:basic_block_overview}). In these variants, the intermediate ReLU is removed to isolate the effect of the tropical convolution operators, which already introduce nonlinearity through min/max operations. The two variants are:


\begin{itemize}
\item TCNN basic block 1: As shown in the Figure ~\ref{images:basic_block_overview}(b), it retains the first ordinary convolution, but a two-parameter compound / parallel tropical convolutional layer replaces the second convolution;
\item TCNN basic block 2: As shown in the Figure ~\ref{images:basic_block_overview}(c), its two convolutions are replaced by two parameters of the compound / parallel tropical convolutional layer;
\end{itemize}


Using the proposed tropical convolutional building blocks, we construct several ResNet variants based on tropical algebra, namely ResNet-Parallel-\Rmnum{1}, ResNet-Parallel-\Rmnum{2}, ResNet-Compound-\Rmnum{1}, and ResNet-Compound-\Rmnum{2}. For comparison, we also evaluate the standard ResNet models provided by the PyTorch library without using pretrained weights. Furthermore, we introduce additional ResNet-Parallel and ResNet-Compound variants that not only replace the residual blocks using the proposed TCNN Basic Block 2, but also replace the first convolutional layer of ResNet with parallel and compound tropical convolutional layers, respectively.

For further comparison, we additionally evaluate an AdderNet-based variant, where standard convolutional layers are replaced with adder layers except for the first convolutional layer. This setting is conceptually similar to the ResNet-Parallel-\Rmnum{2} and ResNet-Compound-\Rmnum{2} variants, which also preserve the first convolutional layer while replacing the remaining convolutional layers with alternative operators. Unlike tropical convolutional layers, however, adder layers do not inherently introduce nonlinear behavior. Therefore, the ReLU activation functions are retained in the AdderNet variant.



\subsubsection{MobileNetV2 Models}

Unlike ResNet, MobileNet is built upon depthwise separable convolutions, which substantially reduce the number of parameters. To construct tropical variants of MobileNetV2, we replace the depthwise convolutional layers with the proposed tropical convolutional layers while keeping the pointwise convolutional layers unchanged. Although pointwise convolutions can also be replaced with tropical convolutional layers, such replacement may not be necessary for two reasons. First, pointwise convolutions use a kernel size of $1 \times 1$, where the spatial max/min operations introduced by tropical convolution become less meaningful. Second, cTCNN and pTCNN introduce additional learnable parameters (e.g., $\alpha$ and $\beta$) compared with standard convolutional layers, which may contradict the lightweight design objective of the MobileNet family when applied to $1 \times 1$ convolutions. Following the approach used in ResNet-Compound-II, we leave the first convolutional layer of MobileNetV2 as standard convolution, while disabling all activation functions within the inverted residual blocks as well as the final ReLU6 activation function.




\subsubsection{Experiment Setup}

\textbf{Data Preprocessing.} For CIFAR-10/100 and SVHN, training images were augmented with random horizontal flipping and random cropping with 4-pixel padding. All images were converted to tensors and normalized using channel-wise means of (0.4914, 0.4822, 0.4465) and standard deviations of (0.2023, 0.1994, 0.2010). The test sets were evaluated under identical normalization without augmentation. For CIFAR, we separate 5,000 images from the training set for validation, while for SVHN, we use the provided test set for evaluation.

For Tiny ImageNet, we employed a different augmentation pipeline. During training, we applied RandomResizedCrop to $64\times64$ with scale range (0.7, 1.0), random horizontal flipping, and AutoAugment with the ImageNet policy, followed by tensor conversion and normalization with mean (0.4802, 0.4481, 0.3975) and standard deviation (0.2302, 0.2265, 0.2262). Test images were converted to tensors and using same normalization, without random cropping or augmentation.

\textbf{ResNet Architectures and Training.} We evaluated two ResNet configurations. The first follows the standard ImageNet-style architecture with a $7\times7$ first convolution, and adds dropout of 0.3. The second is a custom configuration designed for smaller datasets, which replaces the first convolution with a $3\times3$ window size with stride 1, removes the maxpool after the first convolution, and adds dropout (0.2 for CIFAR-10 and 0.3 for CIFAR-100) before the final classification layer. For CIFAR-10/100, we used the Adam optimizer with a learning rate scheduled from $10^{-2}$ to $10^{-6}$ via CosineAnnealingLR, with $T_{max}$ set to the number of epochs, and no weight decay. The batch size was set to 32 for both ResNet-18 and ResNet-34 models. For ResNet, We mainly trained for 50 epochs, except for the custom configuration on CIFAR-100, which was trained for 100 epochs. Tiny ImageNet experiments adopted the ImageNet-style ResNet configuration and were trained for 50 epochs with batch size 32, following the same optimization setup.

\textbf{MobileNetV2 Training.} We similarly adapt two configurations: the ImageNet-style configuration, which uses a $3\times3$ convolution with stride 2, and a custom configuration, which uses a $3\times3$ convolution with stride 1. In both cases, we use the SGD optimizer with a weight decay of $4\times10^{-5}$, an initial learning rate of 0.1, and a learning rate schedule that divides the rate by 10 after 75 and 120 epochs, for a total of 150 epochs.


\textbf{General Remarks.} Unless otherwise specified, all experiments involving other datasets employed the ImageNet-style ResNet configuration with the corresponding dataset-specific preprocessing and hyperparameters.

\subsubsection{ResNet Results}
\label{subsubsection:resnet_results}
Tables~\ref{tab:cifar10_resnet_acc} and~\ref{tab:cifar100_resnet_acc} summarize accuracy on CIFAR-10/100 under two configurations. Overall, the custom configuration consistently improves accuracy across all variants, about 7--12 percent compared to the ImageNet-style setting. 

On CIFAR-10, the best custom-config result is 93.21\% by ResNet34-Compound-\Rmnum{1}, which is 0.63 percent higher than the standard ResNet34 baseline (92.58\%). Similarly, for ResNet18 series, the best custom result is 92.84\% by ResNet18-Compound-\Rmnum{1}, improving over ResNet18 (91.87\%) by 0.97 percent. Furthermore, in ImageNet-style setting, the best accuracy is achieved by ResNet18-Parallel-\Rmnum{2} (85.61\%), which is 1.98 percent higher than ResNet18 (83.63\%). For ResNet34 series, the best accuracy is achieved by ResNet34-Compound-\Rmnum{2} (84.15\%), which is 3.36 percent higher than ResNet34 (80.79\%). 

On CIFAR-100, the best results reach 73.35\% (ResNet34-Parallel-\Rmnum{2}) and 73.32\% (ResNet18-Parallel-\Rmnum{2}) on custom configuration, which are 6.38 and 5.75 percent higher than their corresponding baselines (66.97\% and 67.57\%). This highlights separating maxplus and minplus parameters can potentially improve the performance. On the ImageNet-style setting, the best accuracy is achieved by ResNet18-Compound-\Rmnum{2} (57.72\%), which is 1.02 percent higher than ResNet18 (56.70\%). For ResNet34 series, the best accuracy is achieved by ResNet34-Compound (51.59\%), which is 16.63 percent higher than ResNet34 (34.96\%). 

These results suggest that replacing some standard convolutions with compound/parallel tropical convolutions can improve accuracy in most cases. The effect is configuration- and dataset-dependent; in particular, replacing the first convolution with a tropical operator can be beneficial in some cases (e.g., ImageNet-style ResNet34-Compound on CIFAR-100). In contrast, under ImageNet-style setting, the AdderNet variants consistently achieve lower accuracy than both the standard ResNet baselines and the proposed tropical convolutional variants.

\begin{table}[htbp]
	\centering
	\small
	\setlength{\tabcolsep}{3pt}
	\caption{CIFAR-10 accuracy (\%) for ResNet variants under ImageNet-style and custom configurations. The custom configuration uses a $3\times3$ first convolution and dropout 0.2.}
	\label{tab:cifar10_resnet_acc}
	\begin{tabular}{lcc}
	\toprule
	Model & ImageNet Config & Custom Config \\
	\midrule
	ResNet18-Parallel-\Rmnum{1} & 85.18 & 92.77 \\
	ResNet18-Parallel-\Rmnum{2} & \textbf{85.61} & 92.79 \\
	ResNet18-Parallel & 84.55 & 92.61 \\
	ResNet18-Compound-\Rmnum{1} & 84.96 & \textbf{92.84} \\
	ResNet18-Compound-\Rmnum{2} & 85.17 & 92.54 \\
	ResNet18-Compound & 84.41 & 92.60 \\
	ResNet18 & 83.63 & 91.87 \\
	ResNet18-AdderNet & 74.07 & - \\
	\midrule
	ResNet34-Parallel-\Rmnum{1} & 83.51 & 93.14 \\
	ResNet34-Parallel-\Rmnum{2} & 83.44 & 93.02 \\
	ResNet34-Parallel & 82.05 & 92.85 \\
	ResNet34-Compound-\Rmnum{1} & 83.17 & \textbf{93.21} \\
	ResNet34-Compound-\Rmnum{2} & \textbf{84.15} & 93.10 \\
	ResNet34-Compound & 83.82 & 92.57 \\
	ResNet34 & 80.79 & 92.58 \\
	ResNet34-AdderNet & 73.46 & - \\
	\bottomrule
	\end{tabular}
\end{table}

\begin{table}[htbp]
	\centering
	\small
	\setlength{\tabcolsep}{3pt}
	\caption{CIFAR-100 accuracy (\%) for ResNet variants under ImageNet-style and custom configurations. The custom configuration uses a $3\times3$ first convolution and dropout 0.3.}
	\label{tab:cifar100_resnet_acc}
	\begin{tabular}{lcc}
	\toprule
	Model & ImageNet Config & Custom Config \\
	\midrule
	ResNet18-Parallel-\Rmnum{1} & 56.35 & 73.09 \\
	ResNet18-Parallel-\Rmnum{2} & 57.48 & \textbf{73.32} \\
	ResNet18-Parallel & 56.41 & 72.27 \\
	ResNet18-Compound-\Rmnum{1} & 55.70 & 72.58 \\
	ResNet18-Compound-\Rmnum{2} & \textbf{57.72} & 72.90 \\
	ResNet18-Compound & 56.81 & 71.84 \\
	ResNet18 & 37.08 & 67.57 \\
	ResNet18-AdderNet & 22.45 & - \\
	\midrule
	ResNet34-Parallel-\Rmnum{1} & 49.01 & 73.14 \\
	ResNet34-Parallel-\Rmnum{2} & 49.64 & \textbf{73.35} \\
	ResNet34-Parallel & 49.65 & 72.47 \\
	ResNet34-Compound-\Rmnum{1} & 49.86 & 72.87 \\
	ResNet34-Compound-\Rmnum{2} & 47.36 & 72.59 \\
	ResNet34-Compound & \textbf{51.59} & 72.14 \\
	ResNet34 & 34.96 & 66.97 \\
	ResNet34-AdderNet & 14.85 & - \\
	\bottomrule
	\end{tabular}
\end{table}

For SVHN, we summarize the accuracy results under the earlier training setup in Table~\ref{tab:svhn_resnet_acc}. The best ResNet18 variant achieves 93.73\% (ResNet18-Compound-\Rmnum{1}), which is 1.35 percent above ResNet18 (92.38\%), while the best ResNet34 variant reaches 93.65\% (ResNet34-Parallel-\Rmnum{2}), improving over ResNet34 (92.35\%) by 1.31 percent.

\begin{table}[htbp]
	\centering
	\caption{SVHN accuracy (\%) for ResNet variants under the earlier training setup.}
	\label{tab:svhn_resnet_acc}
	\begin{tabular}{lc}
	\toprule
	Model & Accuracy \\
	\midrule
	ResNet18-Parallel-\Rmnum{1} & 93.59200 \\
	ResNet18-Parallel-\Rmnum{2} & 93.12769 \\
	ResNet18-Compound-\Rmnum{1} & \textbf{93.73079} \\
	ResNet18-Compound-\Rmnum{2} & 93.25446 \\
	ResNet18 & 92.37861 \\
	\midrule
	ResNet34-Parallel-\Rmnum{1} & 93.48878 \\
	ResNet34-Parallel-\Rmnum{2} & \textbf{93.65396} \\
	ResNet34-Compound-\Rmnum{1} & 93.47342 \\
	ResNet34-Compound-\Rmnum{2} & 93.58482 \\
	ResNet34 & 92.34788 \\
	\bottomrule
	\end{tabular}
\end{table}

For Tiny ImageNet, we summarize the accuracy results under the earlier training setup in Table~\ref{tab:tiny_imagenet_resnet_acc_stage1}. The ResNet18-Compound-\Rmnum{1} variant reaches 47.17\%, compared with 39.66\% for the standard ResNet18 under the same setup, i.e., a 7.51-point improvement. It suggests that compound TCNN can potentially improve the performance of ResNet on more complex datasets.

\begin{table}[htbp]
	\centering
	\caption{Tiny ImageNet accuracy (\%) for ResNet variants under the earlier training setup.}
	\label{tab:tiny_imagenet_resnet_acc_stage1}
	\begin{tabular}{lc}
	\toprule
	Model & Accuracy \\
	\midrule
	ResNet18-Compound-\Rmnum{1} & \textbf{47.17} \\
	ResNet18-Compound-\Rmnum{2} & 46.96 \\
	ResNet18-Compound & 47.14 \\
	ResNet18 & 39.66 \\
	\bottomrule
	\end{tabular}
\end{table}

\subsubsection{MobileNetV2 Results}



Table~\ref{tab:cifar10_mobilenet_acc} reports the classification accuracy on CIFAR-10 for MobileNetV2 variants under the previously described training setup. Under the ImageNet-style configuration, the standard MobileNetV2 achieves 82.74\% accuracy, while MobileNetV2-Compound reaches 85.05\%, corresponding to an improvement of 2.31 percentage points. Under the custom configuration, the standard model attains 89.90\% accuracy, and MobileNetV2-Compound achieves 90.24\%, an improvement of 0.34 percentage points. These results indicate that replacing standard depthwise convolutions with their compound counterparts can be beneficial, even with a reduced number of multiplications. 

\begin{table}[htbp]
	\centering
	\small
	\setlength{\tabcolsep}{3pt}
	\caption{CIFAR-10 accuracy (\%) for MobileNet variants under the earlier training setup.}
	\label{tab:cifar10_mobilenet_acc}
	\begin{tabular}{lcc}
	\toprule
	Model & ImageNet Config & Custom Config \\
	\midrule
	MobileNetV2-Compound & \textbf{85.05} & \textbf{90.24}  \\
	MobileNetV2 & 82.74 & 89.90 \\
	\bottomrule
	\end{tabular}
\end{table}



\subsection{cTCNN on Segmentation Task}

\subsubsection{ResNet Backbone}
In this section, we evaluate the performance of the proposed compound tropical convolutional neural network (cTCNN) on semantic segmentation tasks. Experiments are conducted on the CamVid and Pascal VOC datasets using two widely adopted semantic segmentation architectures: a U-Net-style encoder--decoder architecture \citep{ronneberger2015unet} and DeepLabV3 \citep{chen2017deeplabv3}, both equipped with a ResNet18 backbone \citep{he_deep_2016}. The U-Net-based architecture employs skip connections between encoder and decoder stages to recover spatial details, while DeepLabV3 utilizes atrous convolution and Atrous Spatial Pyramid Pooling (ASPP) to capture multi-scale contextual information. In our experiments, the standard ResNet18 backbone in these architectures is replaced with the proposed cTCNN-based ResNet18 variants to evaluate their effectiveness on semantic segmentation tasks. Segmentation performance is evaluated using mean Intersection over Union (mIoU) and compared against standard CNN-based baselines.


Before transfer to downstream segmentation tasks, the backbone networks are first pretrained on Tiny ImageNet using Adam optimization, following the protocol described in the previous subsection, and subsequently refined using SGD. During the second-stage training, the learning rate is initialized to 0.1 and decayed by a factor of 10 at epochs 15 and 25, with training terminated after 35 epochs. The best classification accuracies obtained after the second-stage training are summarized in Table~\ref{tab:tiny_imagenet_resnet_acc_stage2}.

It is important to note that ImageNet-1K pretrained weights are currently unavailable for the proposed tropical convolutional architectures due to the previously discussed implementation and hardware limitations associated with tropical operators. Therefore, instead of relying on random initialization, we perform backbone pretraining on Tiny ImageNet to provide more suitable feature initialization prior to downstream segmentation training.

\begin{table}[htbp]
	\centering
	\caption{Tiny ImageNet accuracy (\%) for ResNet variants after second stage training with SGD.}
	\label{tab:tiny_imagenet_resnet_acc_stage2}
	\begin{tabular}{lc}
	\toprule
	Model & Accuracy \\
	\midrule
	ResNet18-Compound-\Rmnum{1} & \textbf{48.90} \\
	ResNet18-Compound-\Rmnum{2} & 46.99 \\
	ResNet18-Compound & 45.09 \\
	ResNet18 & 42.60 \\
	\bottomrule
	\end{tabular}
\end{table}

\subsubsection{Performance on CamVid Dataset}



In our setup, we use the official CamVid train/validation/test split and follow the common convention of ignoring the ``unlabeled'' category by mapping class id 11 to the ignore label 255 during both training and evaluation. During training, data augmentation includes random horizontal flipping with probability 0.5 and random affine transformations consisting of scaling within the range $[0.8,1.2]$ and translation within $[-0.1,0.1]$, followed by resizing to $384 \times 480$. During validation and testing, images are resized to $384 \times 480$ without additional augmentation. Finally, input images are scaled to the range $[0,1]$ and normalized using the ImageNet mean and standard deviation.

For optimization, we use Adam with cosine annealing learning-rate scheduling (with minimum learning rate $10^{-6}$) and weight decay $10^{-4}$. We train for 100 epochs with batch size 8, and use the sum of cross-entropy loss and Dice loss. The initial learning rate is set to $10^{-2}$ reported in Table~\ref{tab:camvid_unet_iou_lr1e2}. The mIoU is computed by excluding pixels with ignore label 255. We evaluate U-Net with ResNet18 backbone report mIoU for the compound variants and the standard U-Net baseline.

\begin{figure*}
	\centering
    \subfloat[Sample on the CamVid validation split.]{\includegraphics[width=\linewidth]{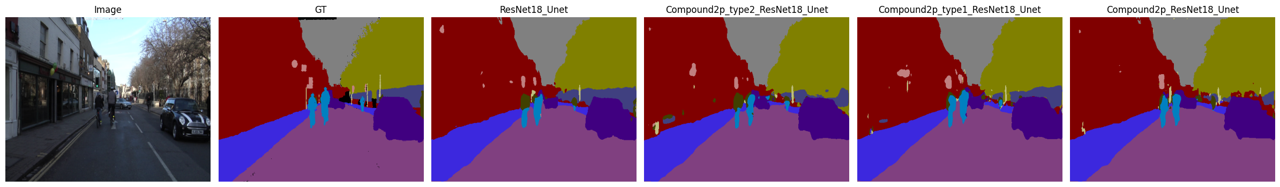}\label{fig:camvid_val_samples}}\\[4pt]
    \subfloat[Sample on the CamVid test split.]{\includegraphics[width=\linewidth]{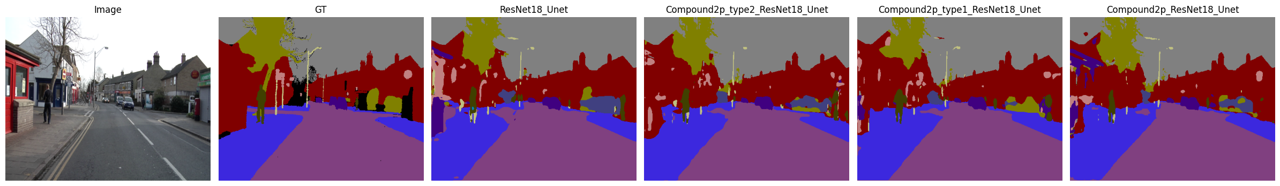}\label{fig:camvid_test_samples}}
    \caption{Segmentation examples on CamVid.}
	\label{images:camvid}
\end{figure*}

On the validation results, the best compound variant (U-Net-ResNet18-Compound) reaches 66.03\% mIoU, improving over the standard U-Net-ResNet18 baseline (64.05\%) by 1.98 percent. The similar trend appears on the test split, where the best compound variant achieves 56.52\% mIoU, improving over the baseline (54.74\%) by 1.78 percent. Sample segmentation results are shown in Figure~\ref{images:camvid}, where the proposed compound variants demonstrate improved segmentation quality compared to the standard U-Net baseline.

\begin{table}[htbp]
    \centering
    \caption{CamVid mean IoU (\%) for U-Net variants with ResNet18 backbone (initial LR $10^{-2}$).}
    \label{tab:camvid_unet_iou_lr1e2}
    \begin{tabular}{lcc}
    \toprule
    Model & val mIoU & test mIoU \\
    \midrule
    U-Net-ResNet18-Compound-\Rmnum{1} & 65.38 & 55.80 \\
    U-Net-ResNet18-Compound-\Rmnum{2} & 63.08 & 55.57\\
    U-Net-ResNet18-Compound & \textbf{66.03} & \textbf{56.52} \\
    U-Net-ResNet18 & 64.05 & 54.74 \\
    \bottomrule
    \end{tabular}
\end{table}

\subsubsection{Performance on Pascal VOC Dataset}



In our setup, the dataset is constructed using the \texttt{VOC\-Segmentation} dataset interface provided by torchvision, with paired image--mask transformations where spatial augmentations are consistently applied to both images and segmentation masks. During training, data augmentation includes random horizontal flipping (\(p=0.5\)), affine transformations with scaling in the range \([0.5,1.5]\) and translation in \([-0.1,0.1]\), perspective transformation, and resizing to \(513 \times 513\). Additional appearance augmentations include Gaussian noise and randomized selection among contrast enhancement operations (CLAHE, random brightness--contrast adjustment, or gamma correction), sharpness/blur operations (sharpening, Gaussian blur, or motion blur), and color perturbations (brightness--contrast adjustment or hue--saturation modification). During validation and testing, only resizing to \(513 \times 513\) is applied. Finally, input images are scaled to the range \([0,1]\) and normalized using the ImageNet mean and standard deviation.

We use Adam with cosine annealing (minimum learning rate $10^{-6}$) and weight decay $10^{-4}$, with initial learning rate $10^{-4}$; we train for 200 epochs with batch size 8, and adopt the combined cross-entropy plus Dice loss. An auxiliary segmentation head is additionally employed during training, where the total loss is computed as the sum of the primary segmentation loss and 0.4 times the auxiliary loss. The mIoU is computed by ignoring pixels with label 255.


For the Pascal VOC dataset, we evaluate DeepLabV3 with a ResNet18 backbone and report the resulting mIoU values in Table~\ref{tab:pascal_voc_deepabv3_iou}. On the validation split, the proposed DeepLabV3-ResNet18-Compound-\Rmnum{1} achieves 47.49\% mIoU compared with 48.70\% for the DeepLabV3-ResNet18 baseline, corresponding to a difference of only 1.21 percentage points. Similarly, on the test split, DeepLabV3-ResNet18-Compound-\Rmnum{1} attains 54.74\% mIoU, remaining within 1.77 percentage points of the baseline result of 56.51\%. Although the standard CNN-based model achieves the highest overall performance, the proposed compound tropical convolutional variants remain competitive under the same training settings, demonstrating that tropical convolutional operators can be effectively integrated into modern semantic segmentation architectures while maintaining comparable segmentation performance.

\begin{table}[htbp]
    \centering
    \small
    \setlength{\tabcolsep}{3pt}
    \caption{Pascal VOC mean IoU (\%) for DeepLabV3 variants with ResNet18 backbone.}
    \label{tab:pascal_voc_deepabv3_iou}
    \begin{tabular}{lcc}
    \toprule
    Model & val mIoU & test mIoU \\
    \midrule
    DeepLabV3-ResNet18-Compound-\Rmnum{1} & 47.49 & 54.74 \\
    DeepLabV3-ResNet18-Compound-\Rmnum{2} & 45.03 & 51.93 \\
    DeepLabV3-ResNet18 & \textbf{48.70} & \textbf{56.51} \\
    \bottomrule
    \end{tabular}
\end{table}


\section{Conclusion}\label{sec:conclusion}
In conclusion, this paper presents a tropical algebra-based framework for convolutional neural networks that extends conventional CNN architectures with MaxPlus and MinPlus operators. We designed and implemented a hierarchical tropical CNN framework and evaluated it on multiple classification and semantic segmentation benchmarks. The proposed Compound Tropical Convolutional Neural Networks (cTCNN) and Parallel Tropical Convolutional Neural Networks (pTCNN) show that tropical-algebra operators can serve as practical alternatives to the standard convolutional layer within a modern CNN block, and can achieve competitive accuracy with reduced reliance on multiplication-intensive computations.

By integrating tropical convolutional blocks into deeper architectures such as ResNet and MobileNetV2, the proposed models can achieve competitive performance across multiple datasets. Furthermore, experiments on CamVid and Pascal VOC using U-Net-style and DeepLabV3-based architectures with ResNet18 backbones indicate that cTCNN-based variants remain effective for semantic segmentation under standard training settings. These results suggest that tropical convolutional operators are a viable option when exploring alternative operator designs for deep neural networks.

In addition, this work presents a practical GPU-oriented implementation of tropical convolutional operators using TileLang integrated within a PyTorch-based framework. Although current GPU architectures provide dedicated Tensor Core acceleration primarily for conventional multiply--accumulate (MAC) operations, no comparable hardware support currently exists for MaxPlus and MinPlus computations. Consequently, the full computational advantages of tropical convolution cannot yet be fully realized on existing hardware platforms. Nevertheless, the proposed implementation shows that tropical convolutional networks can be integrated into modern deep learning pipelines and used for practical experimentation.

Overall, the proposed framework suggests that tropical convolution-based neural networks can offer a useful trade-off between representational capacity and computational cost, particularly in resource-constrained settings. Future work will focus on hardware-aware optimization, dedicated MaxPlus/MinPlus acceleration, and further evaluation of tropical convolutional networks in larger-scale and specialized tasks.

\appendices
\section{TCNN Framework}\label{appendix:tcnn_framework}

\begin{figure}
    \centering
    \includegraphics[width=\linewidth]{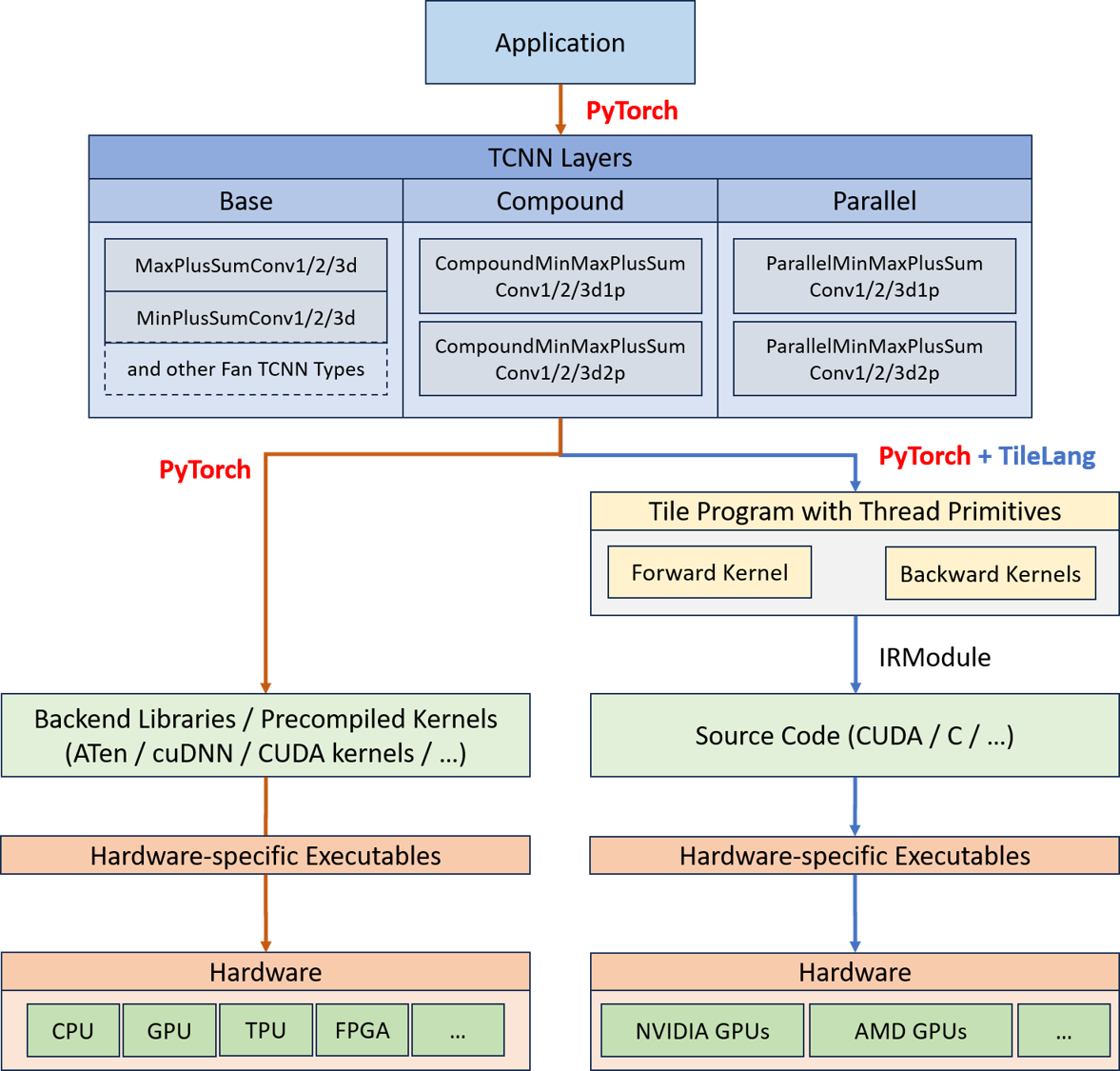}
    \caption{TCNN Framework}
    \label{fig:tcnnframework}
\end{figure}

Here, we detail the TCNN framework we have implemented as Figure \ref{fig:tcnnframework}.

\subsection{Layer Design}
\subsubsection{Implementation Without TileLang}

This implementation is based on PyTorch, and it realizes the TCNN operators using PyTorch's automatic differentiation mechanisms. Since relying on PyTorch library for the entire implementation, this mode is easy to implement, and can be used throughout different hardware, including CPU, GPU, and other PyTorch supported hardware.

The PyTorch-based TCNN layer primarily executes forward computation in three steps: 1) preprocessing, 2) computational operations, and 3) channel operations. All steps are relying on PyTorch library for lower implementation, and require to save the intermediate results for backward propagation. The preprocessing step involves operations such as unfolding the input feature map. The computational operations step includes the core convolutional computations, such as max, min, and plus operations. Finally, the channel operations step involves combining the outputs from different branches (e.g., min-plus and max-plus) using learnable parameters like $\alpha$ and $\beta$, which are specific to each layer variant. 


\subsubsection{Implementation With TileLang}

Unlike the implementation without TileLang, the implementation with TileLang directly realizes tropical convolution operators using the TileLang domain-specific language, thereby providing greater flexibility for low-level performance optimization. In contrast, the implementation without TileLang primarily relies on PyTorch to organize the forward and backward propagation processes. However, for convolutional operators with relatively complex internal computation patterns, relying entirely on PyTorch automatic differentiation mechanisms is not always efficient.

As illustrated in Figure~\ref{fig:tcnnframework}, the implementation with TileLang maintains a similar high-level usage interface to the implementation without TileLang, where convolutional layers and operators can still be invoked through standard PyTorch interfaces. During training or inference, the framework automatically calls the forward and backward propagation kernels implemented using TileLang. These kernels are first represented in an Intermediate Representation (IR) to encode their computational semantics. Subsequently, the compilation system generates hardware-specific source code (e.g., CUDA code) according to the target platform and compiles it at runtime into executable binary programs for GPU execution. Through this design, the implementation with TileLang preserves the usability and flexibility of the high-level PyTorch interface while enabling specialized low-level optimizations for TCNN operators.

\subsection{Compatibility}

When designing TCNN, we prioritized seamless integration with the existing AI ecosystem rather than creating an isolated system. To this end, TCNN is built as a robust enhancement to current frameworks, respecting established tools and methodologies. This compatibility minimizes the learning curve for new adopters and enables existing applications to be migrated with minimal effort and without extensive reengineering. Consequently, TCNN offers a smooth transition and a powerful extension to the current AI landscape.

\subsubsection{Compatibility with PyTorch}
We designed our convolution operation as an optimized alternative to the standard convolution operation, ensuring it fully respects the usage habits of developers and users. This means that, as illustrated in the code block, our parameter list mirrors that of traditional PyTorch-based convolution calls. In fact, transitioning to our framework can be as simple as changing the function name, as demonstrated in Listing ~\ref{lst:use_tcnn}.


\begin{lstlisting}[caption={when using a tropical convolutional neural network,  you don't need to change your habits}, label={lst:use_tcnn}]
import torch.nn as nn
import tcnn.layers as tlayers
# To use TileLang-kernel-based layers, use tl_tcnn
# import tl_tcnn as tlayers
# The way you defined convolutional neural networks before
conv = nn.Conv2d(3, 6, kernel_size=5, padding=2)
# Now you define the way the tropical convolutional layer is
conv = tlayers.MinPlusSumConv2d(3, 6, kernel_size=5, padding=2)
conv = tlayers.MaxPlusSumConv2d(3, 6, kernel_size=5, padding=2)
# Now you define the way the tropical compound convolutional layer is
##  compound convolutional layer for one parameter
conv = tlayers.CompoundMinMaxPlusSumConv2d1p(3, 6, kernel_size=5, padding=2)
##  compound convolutional layer for two parameters
conv = tlayers.CompoundMinMaxPlusSumConv2d2p(3, 6, kernel_size=5, padding=2)
# Now you define the way the tropical parallel convolutional layer is
##  parallel convolutional layer for one parameter
conv = tlayers.ParallelMinMaxPlusSumConv2d1p(3, 6, kernel_size=5, padding=2)
##  parallel convolutional layer for two parameters
conv = tlayers.ParallelMinMaxPlusSumConv2d2p(3, 6, kernel_size=5, padding=2)
\end{lstlisting}

Moreover, drawing from our engineering insights, we ensured that our operators are invoked on classes that directly or indirectly inherit from PyTorch's \_Conv. This design choice allows networks built on our tropical convolutional neural network to use the same parameter calculation methods as the original, ensuring full compatibility with existing interfaces. As a result, when functions like print or torchsummary.summary are called, our code integrates seamlessly, behaving just as traditional PyTorch layers would.

Furthermore, this compatibility extends to GPU acceleration. Our tropical convolution layer naturally supports GPU execution, significantly enhancing code efficiency—especially crucial in the era of large models where expanding the parameter scale is essential. Importantly, our tropical convolutional layers can also be mixed with traditional convolutional layers within the same model, enabling both methods to complement each other and maximize model performance. This flexibility allows developers to harness the unique strengths of each approach, ultimately leading to more powerful and efficient neural networks.

\subsubsection{Compatibility with PyTorch-based frameworks}
We have ensured that the tropical convolutional neural network is not only highly compatible with the existing PyTorch ecosystem but also naturally adaptable to other application frameworks developed on PyTorch, such as SecretFlow~\cite{secretflow} and NNI~\cite{nni2021}. This broad compatibility allows for seamless integration and utilization of TCNN within a wide range of AI development environments, enabling developers to leverage TCNN's benefits across multiple platforms with minimal adjustments.

\section{Extensive Experiments}

\subsection{Ablation Study of cTCNN's parameter importance on MobileNet}\label{subsec:ablationStudyParametersOnMobilenet}

This section presents an ablation study to investigate the importance of the learnable parameters $\alpha$ and $\beta$ in the tropical convolutional layers on the MobileNetV2 architecture. Unlike the experiment setup introduced in Section~\ref{subsec:deepnet}, we use a slightly different training configuration for the MobileNetV2 experiments. Specifically, we train the models for 200 epochs using SGD with an initial learning rate of 0.1, which is reduced by a factor of 0.1 at epochs 100 and 150. Other settings remain the same as in Section~\ref{subsec:deepnet}.

\begin{figure*}
	\centering
	\includegraphics[width=\linewidth]{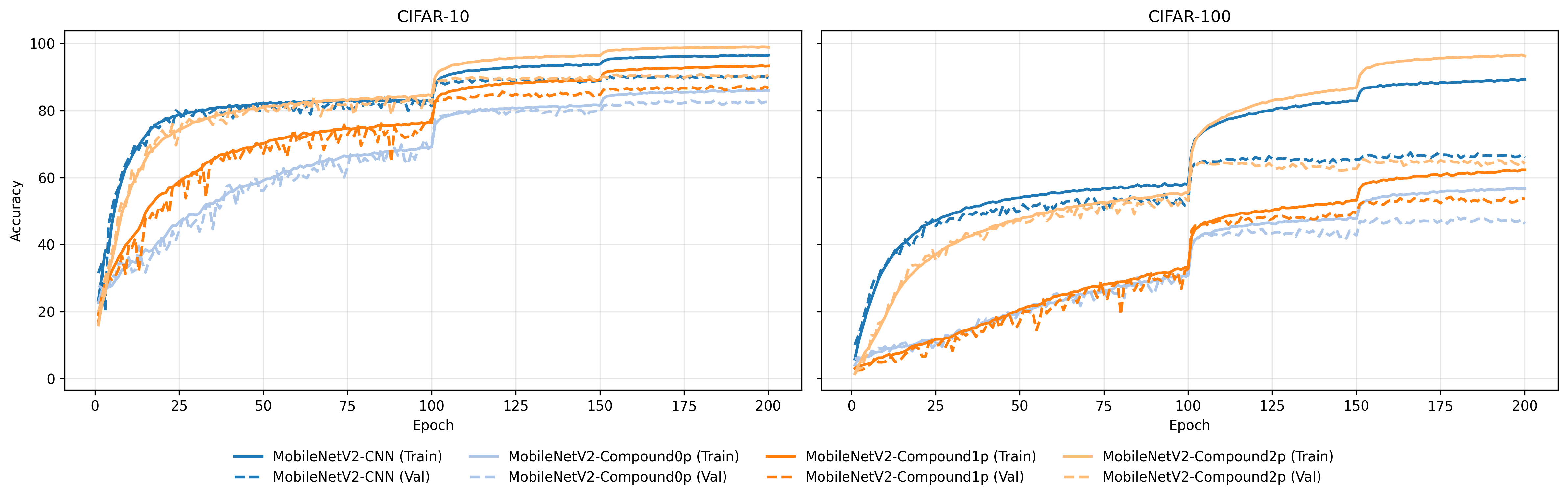}
	\caption{MobileNetV2 Variants Training and Validation Accuracy Comparison on CIFAR datasets}
	\label{images:MobileNetV2_CIFAR_Train_Val_Accuracy_comparison}
\end{figure*}

Figure~\ref{images:MobileNetV2_CIFAR_Train_Val_Accuracy_comparison} presents the training and validation accuracy curves for various MobileNetV2 variants on the CIFAR-10 and CIFAR-100 datasets. The results reveal that the removal of multiplicative operations from the tropical convolution (MobileNetV2-Compound0p) leads to a substantial degradation in accuracy relative to the conventional CNN baseline. Introducing a single learnable parameter, $\alpha$ (MobileNetV2-Compound1p), yields moderate improvements over the Compound0p variant, though it still underperforms the standard model. In contrast, the Compound2p variant, which incorporates both $\alpha$ and $\beta$ as learnable parameters for the minplus and maxplus operations respectively, not only surpasses the other tropical variants but also achieves performance comparable to the conventional CNN-based counterpart. Collectively, these findings underscore the importance of the dual learnable parameterization in preserving the representational capacity of tropical convolutional layers within the MobileNetV2 framework.

\subsection{One-dimensional data classification}\label{sec:1dc-TCNN}
\subsubsection{Dataset}
The UrbanSound8K~\citep{Salamon:UrbanSound:ACMMM:14} dataset is a widely used dataset for audio classification tasks. It consists of 8732 labeled sound excerpts 
from urban environments, with a duration of 4 seconds each. The dataset covers 10 different classes, including air conditioner, 
car horn, children playing, dog bark, drilling, engine idling, gun shot, jackhammer, siren, and street music. 
The dataset is diverse and challenging, with a wide range of acoustic characteristics and background noise. 
It is commonly used for tasks such as sound event detection, audio tagging, and urban sound classification. 
The UrbanSound8K dataset provides a valuable resource for researchers and practitioners working in the field of audio analysis and 
machine learning.

The Speech Command dataset~\citep{warden2018speech} is a widely used dataset for speech recognition tasks. It consists of a large number of short audio clips, each containing a spoken word or command. The dataset covers a wide range of commands, such as "yes", "no", "up", "down", "left", "right", "stop", "go", and many others.The Speech Command dataset is commonly used to train and evaluate machine learning models for speech recognition and keyword spotting. It provides a valuable resource for researchers and practitioners working in the field of audio analysis and natural language processing.


The ECG Heartbeat Categorization Dataset is from Kaggle~\citep{kaggle_ecg} and as its abstract introduction describes:this dataset is composed of two collections of heartbeat signals derived from two famous datasets in heartbeat classification, the MIT-BIH Arrhythmia Dataset ~\citep{moody2001impact,goldberger2000physiobank} and The PTB Diagnostic ECG Database~\citep{bousseljot1995nutzung, goldberger2000physiobank}. For more detail, please refer to \url{https://www.kaggle.com/datasets/shayanfazeli/heartbeat}.

\begin{figure}
	\centering
	\includegraphics[scale=0.18]{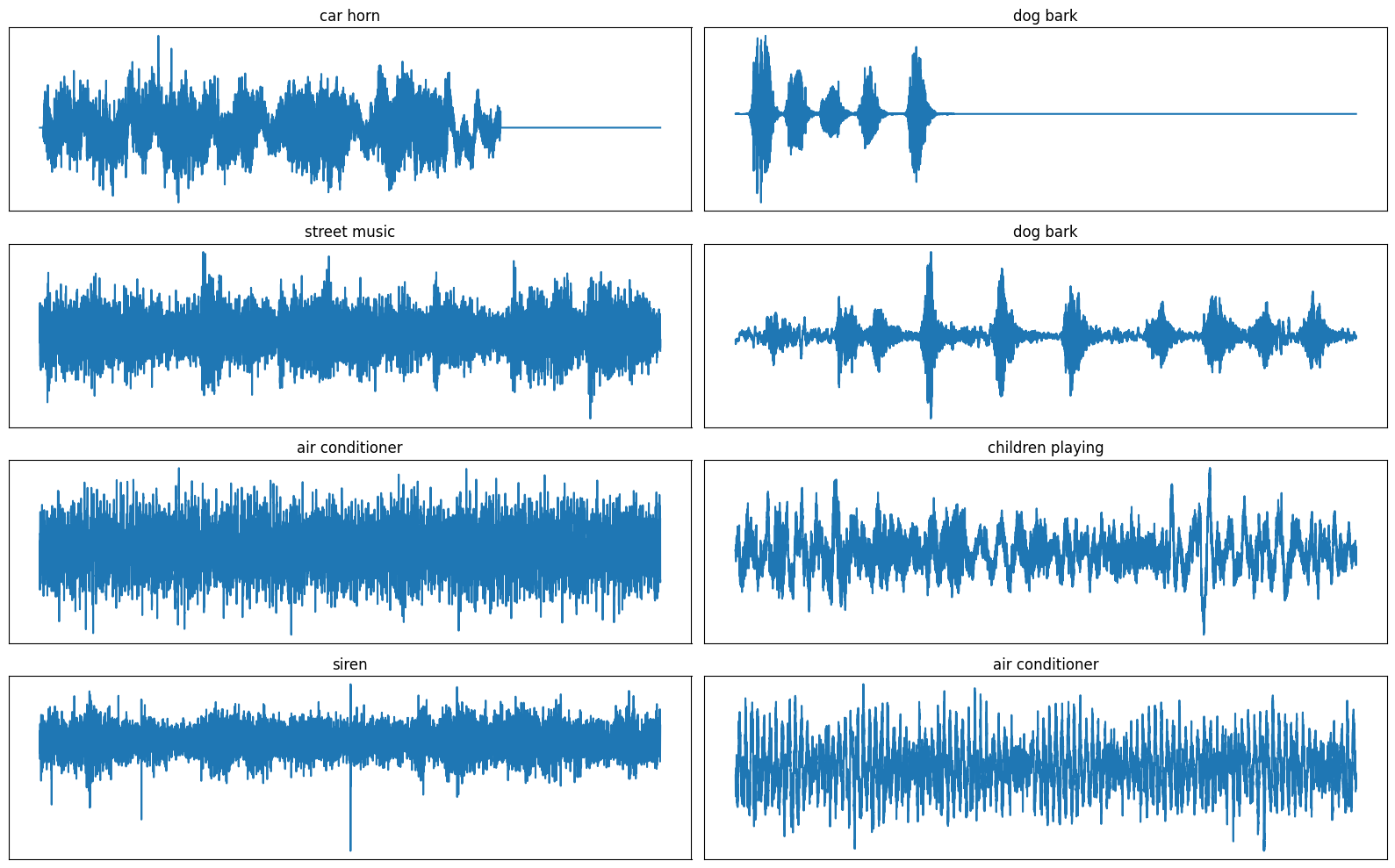}
	\caption{Samples of the UrbanSound8K dataset}
	\label{images:urban8k_sample}
\end{figure}

We took some samples of the UrbanSound8K dataset and visualized them to show the 1D data, as shown in the Figure ~\ref{images:urban8k_sample}.

\subsubsection{Experiments Setup}
To evaluate the effectiveness of the compound tropical convolutional layers in one-dimensional classification tasks, we select the LeNet as the baseline. we alternative the convolutional layers in LeNet with the tropical convolutional layer, the compound tropical convolutional layers, the parallel tropical convolutional layer, and then we name this models as LeNet-Tropical, LeNet-Compound, LeNet-Parallel, respectively. What's more, we alternative the activation function in LeNet with the ReLU function, and then we name this model as LeNet-ReLU. Refer to ~\cite{dai2017very}, we set the size of the kernel of the first layer is 80 and the size of  the kernel of the second layer is 3, which could extract the feature from the input effectively. The architecture of the models is shown as the table ~\ref{tab:overall-architecture}. We train and evaluate these models on the UrbanSound8K dataset, Speech Command dataset, and ECG Heartbeat Categorization Dataset. We set batch size as 32, learning rate as 0.001, and the number of epochs as 100. We use the Adam optimizer with initial learning rate 0.001 to optimize the models. For the epochs setting, based on our observation of the convergence of the model for a single comparison experiment, we selected the appropriate settings for each dataset, so the epochs of the UrbanSound8K dataset were set to 200; The epochs for the Speech Command dataset are set to 100 and the epochs for the ECG Heartbeat Categorization dataset are set to 100. To compare models fairly, we use the same training and testing data for all models and repeat all experiments five times with the same configuration for all models. Finaly, we report the mean and the standard deviation of the accuracy, AUC, F1, precision, and recall of the models.

\subsubsection{Results}
The experimental results presented in Table ~\ref{tab:1d-TCNN-results} provide a comprehensive evaluation of various LeNet-based models applied to three distinct datasets: UrbanSound8K, Speech Command, and ECG Heartbeat Categorization. Each model's performance was assessed using multiple metrics, including accuracy, AUC (Area Under the Curve), F1 score, precision, recall, $\Omega_{u}$, and the number of parameters.

In the UrbanSound8K dataset, the LeNet-ReLU model demonstrates superior performance, achieving the highest accuracy (88.56\%) and recall (0.89) among the evaluated models. This suggests its robustness in handling the complexities inherent in auditory classification tasks. However, this performance comes at the cost of increased computational demands, with a total of 19.92 $\times 10^8$ operations ($\Omega_{u}$) and 15,333,524 parameters required. In contrast, the LeNet-CM model, whose convolution layers consist of standard convolution and compound tropical convolution, delivers a slightly lower accuracy (87.04\%) but does so with a significantly reduced computational load (12.56 $\times 10^8$ $\Omega_{u}$). This highlights a more favorable balance between efficiency and accuracy, making it a strong contender for scenarios where computational resources are limited. Besides, LeNet-F-\Rmnum{2} also achieves comparable performance while requiring less computation(5.02 $\times 10^8$ $\Omega_{u}$). Other models such as LeNet-P-$\alpha$, while computationally efficient, exhibit substantially lower accuracy (54.93\%) and F1 score (0.57), indicating that they may be insufficient for tasks requiring detailed feature extraction.

The Speech Command dataset presents a more challenging classification task, which is a dataset with 35 categories, as reflected by the generally lower performance across all models. The LeNet-P-$\alpha/\beta$ again stands out with the highest accuracy (46.71\%), AUC (0.93), and F1 score (0.46) among the tested models. Despite the modest accuracy, the LeNet-C-$\alpha / \beta$ model offers a compelling trade-off between computational cost and performance, operating with 1.66 $\times 10^8$ $\Omega_{u}$ and 3,815,853 parameters. In contrast, the LeNet-ReLU model, while computationally the most demanding with 4.94 $\times 10^8$ $\Omega_{u}$, does not deliver a corresponding increase in accuracy, achieving only 43.37\%. This suggests that convolutional neural networks based on tropical algebra can not only reduce the amount of computation, but also meet or even exceed the performance of standard convolutional networks.

In the ECG Heartbeat Categorization dataset, the LeNet-C models, particularly the LeNet-C-$\alpha/\beta$ variants, demonstrate exceptional performance, achieving accuracy as high as 98.18\% with corresponding high precision and recall metrics. These models are also notably efficient, requiring as few as 2.44 $\times 10^6$ operations ($\Omega_{u}$) and 63,441 parameters, indicating their suitability for tasks where computational resources are a concern. The overall high performance across models for this dataset suggests that it is inherently more predictable, with most models achieving accuracies above 97\%. Despite its higher computational cost (6.98 $\times 10^6$ Ops$_{u}$), the LeNet-ReLU model does not offer a significant performance advantage, further supporting the efficacy of the simpler LeNet-PM variants.

Overrall, in the face of different types of one-dimensional data classification tasks, neural networks based on tropical convolutions can always reduce the computational cost to a certain extent due to their design, and at the same time, these models can meet or even surpass the methods based on standard convolution, which not only shows that these models can achieve a computational-performance balance, but also can achieve better linear fitting ability.

\begin{table*}[htbp]
	\centering
	\caption{Performance comparison across different models and datasets. The table presents the accuracy, AUC (Area Under the Curve), F1 score, precision, recall, number of operations ($\Omega_{u}$ with $\theta = 10$), and the number of parameters for various LeNet-based models tested on UrbanSound8K, Speech Command, and ECG Heartbeat Categorization datasets.}
	\begingroup
	\setlength{\tabcolsep}{3pt}
	\renewcommand{\arraystretch}{0.95}
	\scriptsize
	\resizebox{\textwidth}{!}{%
	\begin{tabular}{llccccccrl}
	\toprule
	Dataset & Model & Accuracy &Auc&F1&Precision&Recall& $\Omega_{u}$($\theta$ = 10) & Parameter \\

	\midrule
	UrbanSound8K & LeNet-P-$\alpha$ & 54.93 $ \pm $ 18.65 & 0.85 $ \pm $ 0.09 & 0.57 $ \pm $ 0.18 & 0.71 $ \pm $ 0.12 & 0.57 $ \pm $ 0.17 & \num{8.33e8} & 15334394 \\
&LeNet-P-$\alpha/ \beta$ & 77.11 $ \pm $ 5.29 & 0.95 $ \pm $ 0.02 & 0.78 $ \pm $ 0.05 & 0.81 $ \pm $ 0.03 & 0.78 $ \pm $ 0.05  & \num{8.33e8} & 15334496 \\
&LeNet-PM-$\alpha$ & 71.84 $ \pm $ 30.51 & 0.88 $ \pm $ 0.19 & 0.70 $ \pm $ 0.34 & 0.70 $ \pm $ 0.35 & 0.72 $ \pm $ 0.31 & \num{13.48e8} & 15334010 \\
&LeNet-PM-$\alpha/ \beta$  & 86.40 $ \pm $ 0.42 & 0.97 $ \pm $ 0.00 & 0.87 $ \pm $ 0.00 & 0.87 $ \pm $ 0.01 & 0.87 $ \pm $ 0.00  & \num{13.48e8} & 15334016 \\
&LeNet-C-$\alpha$ & 57.27 $ \pm $ 23.38 & 0.84 $ \pm $ 0.10 & 0.60 $ \pm $ 0.21 & 0.82 $ \pm $ 0.04 & 0.60 $ \pm $ 0.22  & \num{6.68e8} & 15333626 \\
&LeNet-C-$\alpha / \beta$ & 73.26 $ \pm $ 17.85 & 0.92 $ \pm $ 0.07 & 0.75 $ \pm $ 0.16 & 0.84 $ \pm $ 0.05 & 0.75 $ \pm $ 0.16 & \num{6.68e8}  & 15333728 \\
& LeNet-CM-$\alpha$ & 87.04 $ \pm $ 0.37 & 0.98 $ \pm $ 0.00 & 0.87 $ \pm $ 0.00 & 0.88 $ \pm $ 0.00 & 0.87 $ \pm $ 0.00 & \num{12.56e8}  & 15333530 \\
&LeNet-CM-$\alpha / \beta$ & 87.07 $ \pm $ 0.36 & 0.98 $ \pm $ 0.00 & 0.87 $ \pm $ 0.00 & 0.87 $ \pm $ 0.00 & 0.87 $ \pm $ 0.00& \num{12.56e8} & 15333536  \\
&LeNet-F-\Rmnum{1} & 31.10 $ \pm $ 4.58 & 0.73 $ \pm $ 0.04 & 0.32 $ \pm $ 0.04 & 0.55 $ \pm $ 0.08 & 0.35 $ \pm $ 0.04 & \num{5.02e8} & 15333524 \\
&LeNet-F-\Rmnum{2} & 87.40 $ \pm $ 1.99 & 0.97 $ \pm $ 0.01 & 0.87 $ \pm $ 0.02 & 0.87 $ \pm $ 0.02 & 0.88 $ \pm $ 0.02 & \num{5.02e8} & 15333524 \\
&LeNet-F-\Rmnum{3} & 86.93 $ \pm $ 0.61 & 0.98 $ \pm $ 0.00 & 0.87 $ \pm $ 0.01 & 0.88 $ \pm $ 0.00 & 0.87 $ \pm $ 0.01 & \num{11.64e8} & 15333524 \\
&LeNet & 10.83 $ \pm $ 0.00 & 0.50 $ \pm $ 0.00 & 0.02 $ \pm $ 0.00 & 0.01 $ \pm $ 0.00 & 0.10 $ \pm $ 0.00  & \num{19.96e8} & 15333524 \\
&LeNet-ReLU & \textbf{88.56 $ \pm $ 0.62} & 0.98 $ \pm $ 0.00 & 0.89 $ \pm $ 0.01 & 0.89 $ \pm $ 0.01 & 0.89 $ \pm $ 0.01  & \num{19.92e8} & 15333524 \\
	\midrule
\multirow{2}{*}{\begin{tabular}{@{}l@{}}Speech \\ Command\end{tabular}}  &LeNet-P-$\alpha$ & 37.50 $ \pm $ 7.31 & 0.89 $ \pm $ 0.03 & 0.36 $ \pm $ 0.08 & 0.37 $ \pm $ 0.07 & 0.36 $ \pm $ 0.07 & \num{2.07e8} & 3816519\\
&LeNet-P-$\alpha/ \beta$ & \textbf{46.70 $ \pm $ 1.18} & 0.93 $ \pm $ 0.00 & 0.46 $ \pm $ 0.01 & 0.46 $ \pm $ 0.01 & 0.46 $ \pm $ 0.01 & \num{2.07e8}  & 3816621\\
&LeNet-PM-$\alpha$ & 45.01 $ \pm $ 1.02 & 0.92 $ \pm $ 0.00 & 0.44 $ \pm $ 0.01 & 0.44 $ \pm $ 0.01 & 0.44 $ \pm $ 0.01 & \num{3.35e8}  & 3816135\\
&LeNet-PM-$\alpha/ \beta$ & 44.40 $ \pm $ 0.68 & 0.92 $ \pm $ 0.00 & 0.43 $ \pm $ 0.01 & 0.43 $ \pm $ 0.01 & 0.43 $ \pm $ 0.01 & \num{3.35e8}  & 3816141\\
&LeNet-C-$\alpha$ & 36.37 $ \pm $ 7.26 & 0.89 $ \pm $ 0.03 & 0.34 $ \pm $ 0.08 & 0.35 $ \pm $ 0.08 & 0.35 $ \pm $ 0.07 & \num{1.66e8}  & 3815751\\
&LeNet-C-$\alpha/\beta$ & 45.91 $ \pm $ 1.31 & 0.93 $ \pm $ 0.00 & 0.45 $ \pm $ 0.01 & 0.45 $ \pm $ 0.01 & 0.45 $ \pm $ 0.01 & \num{1.66e8} & 3815853\\
&LeNet-CM-$\alpha$ & 45.86 $ \pm $ 1.28 & 0.93 $ \pm $ 0.00 & 0.45 $ \pm $ 0.01 & 0.45 $ \pm $ 0.01 & 0.44 $ \pm $ 0.01 & \num{3.12e8} & 3815655\\
&LeNet-CM-$\alpha / \beta$ & 42.36 $ \pm $ 0.81 & 0.91 $ \pm $ 0.00 & 0.41 $ \pm $ 0.01 & 0.42 $ \pm $ 0.01 & 0.41 $ \pm $ 0.01 & \num{3.12e8} & 3815661\\
&LeNet-F-\Rmnum{1} & 22.91 $ \pm $ 0.84 & 0.83 $ \pm $ 0.01 & 0.20 $ \pm $ 0.01 & 0.22 $ \pm $ 0.01 & 0.21 $ \pm $ 0.01 & \num{1.25e8} & 3815649\\
&LeNet-F-\Rmnum{2} & 31.99 $ \pm $ 2.41 & 0.88 $ \pm $ 0.01 & 0.29 $ \pm $ 0.02 & 0.30 $ \pm $ 0.02 & 0.30 $ \pm $ 0.02 & \num{1.25e8} & 3815649\\
&LeNet-F-\Rmnum{3} & 41.81 $ \pm $ 1.21 & 0.92 $ \pm $ 0.00 & 0.40 $ \pm $ 0.01 & 0.41 $ \pm $ 0.01 & 0.40 $ \pm $ 0.01  & \num{2.89e8} & 3815649\\
&LeNet & 3.80 $ \pm $ 0.00 & 0.50 $ \pm $ 0.00 & 0.00 $ \pm $ 0.00 & 0.00 $ \pm $ 0.00 & 0.03 $ \pm $ 0.00 & \num{4.96e8} & 3815649\\
&LeNet-ReLU & 43.37 $ \pm $ 3.79 & 0.91 $ \pm $ 0.01 & 0.42 $ \pm $ 0.04 & 0.42 $ \pm $ 0.04 & 0.42 $ \pm $ 0.04 & \num{4.94e8} & 3815649 \\
	\midrule
    
\multirow{2}{*}{\begin{tabular}{@{}l@{}}ECG \\ Heartbeat \\ Categorization \end{tabular}}  
&LeNet-P-$\alpha$  & 97.89 $ \pm $ 0.12 & 0.98 $ \pm $ 0.00 & 0.89 $ \pm $ 0.00 & 0.90 $ \pm $ 0.00 & 0.88 $ \pm $ 0.01 & \num{3.01e6} & 64209\\
&LeNet-P-$\alpha/ \beta$ & 98.07 $ \pm $ 0.05 & 0.98 $ \pm $ 0.00 & 0.90 $ \pm $ 0.00 & 0.92 $ \pm $ 0.01 & 0.88 $ \pm $ 0.01 & \num{3.01e6} & 64311\\
&LeNet-PM-$\alpha$ & 97.97 $ \pm $ 0.09 & 0.98 $ \pm $ 0.00 & 0.90 $ \pm $ 0.01 & 0.91 $ \pm $ 0.02 & 0.88 $ \pm $ 0.01 & \num{4.72e6} & 63825 \\
&LeNet-PM-$\alpha/ \beta$ & 98.11 $ \pm $ 0.06 & 0.98 $ \pm $ 0.00 & 0.90 $ \pm $ 0.00 & 0.92 $ \pm $ 0.01 & 0.89 $ \pm $ 0.00 & \num{4.72e6} & 63831\\
&LeNet-C-$\alpha$ & 98.03 $ \pm $ 0.08 & 0.98 $ \pm $ 0.00 & 0.90 $ \pm $ 0.00 & 0.92 $ \pm $ 0.01 & 0.88 $ \pm $ 0.01 & \num{2.44e6} & 63441 \\
&LeNet-C-$\alpha / \beta$ & \textbf{98.18 $ \pm $ 0.04} & 0.98 $ \pm $ 0.00 & 0.90 $ \pm $ 0.00 & 0.92 $ \pm $ 0.01 & 0.89 $ \pm $ 0.01  & \num{2.44e6} & 63543 \\
&LeNet-CM-$\alpha$ & 98.03 $ \pm $ 0.13 & 0.98 $ \pm $ 0.00 & 0.90 $ \pm $ 0.01 & 0.92 $ \pm $ 0.01 & 0.88 $ \pm $ 0.01  & \num{4.40e6} & 63345 \\
&LeNet-CM-$\alpha / \beta$ & 98.01 $ \pm $ 0.08 & 0.98 $ \pm $ 0.00 & 0.90 $ \pm $ 0.01 & 0.92 $ \pm $ 0.01 & 0.88 $ \pm $ 0.01 & \num{4.40e6} & 63351 \\
&LeNet-F-\Rmnum{1} & 96.74 $ \pm $ 0.14 & 0.98 $ \pm $ 0.00 & 0.84 $ \pm $ 0.01 & 0.89 $ \pm $ 0.00 & 0.80 $ \pm $ 0.01 & \num{1.84e6} & 63339\\
&LeNet-F-\Rmnum{2} & 97.96 $ \pm $ 0.05 & 0.99 $ \pm $ 0.00 & 0.90 $ \pm $ 0.00 & 0.91 $ \pm $ 0.01 & 0.88 $ \pm $ 0.01 & \num{1.84e6} & 63339\\
&LeNet-F-\Rmnum{3} & 97.83 $ \pm $ 0.10 & 0.98 $ \pm $ 0.00 & 0.89 $ \pm $ 0.01 & 0.91 $ \pm $ 0.01 & 0.87 $ \pm $ 0.01 & \num{4.08e6} & 63339\\
&LeNet & 97.89 $ \pm $ 0.09 & 0.99 $ \pm $ 0.00 & 0.90 $ \pm $ 0.00 & 0.92 $ \pm $ 0.01 & 0.88 $ \pm $ 0.02 & \num{7.00e6} & 63339 \\
&LeNet-ReLU & 98.10 $ \pm $ 0.11 & 0.98 $ \pm $ 0.00 & 0.90 $ \pm $ 0.00 & 0.92 $ \pm $ 0.01 & 0.89 $ \pm $ 0.00 & \num{6.98e6} & 63339 \\
	\bottomrule
	\end{tabular}%
	}
	\endgroup
	\label{tab:1d-TCNN-results}
\end{table*}

\subsection{Three-dimensional data classification }\label{sec:3dc-TCNN}
\subsubsection{Dataset}
MedMNIST v2 is a comprehensive collection of standardized biomedical image datasets, resembling the MNIST dataset. It includes 12 datasets for 2D images and 6 for 3D images. Each image is pre-processed to a small size of 28×28 pixels for 2D or 28×28×28 for 3D, with corresponding classification labels, making it user-friendly without requiring prior knowledge. Covering key biomedical image modalities, MedMNIST v2 is designed for classification tasks on lightweight 2D and 3D images across various dataset sizes (ranging from 100 to 100,000 images) and tasks (binary/multi-class classification, ordinal regression, and multi-label classification). The dataset comprises a total of 708,069 2D images and 9,998 3D images, supporting a wide range of research and educational applications in biomedical image analysis, computer vision, and machine learning.

\begin{figure*}
	\centering
	\includegraphics[scale=0.24]{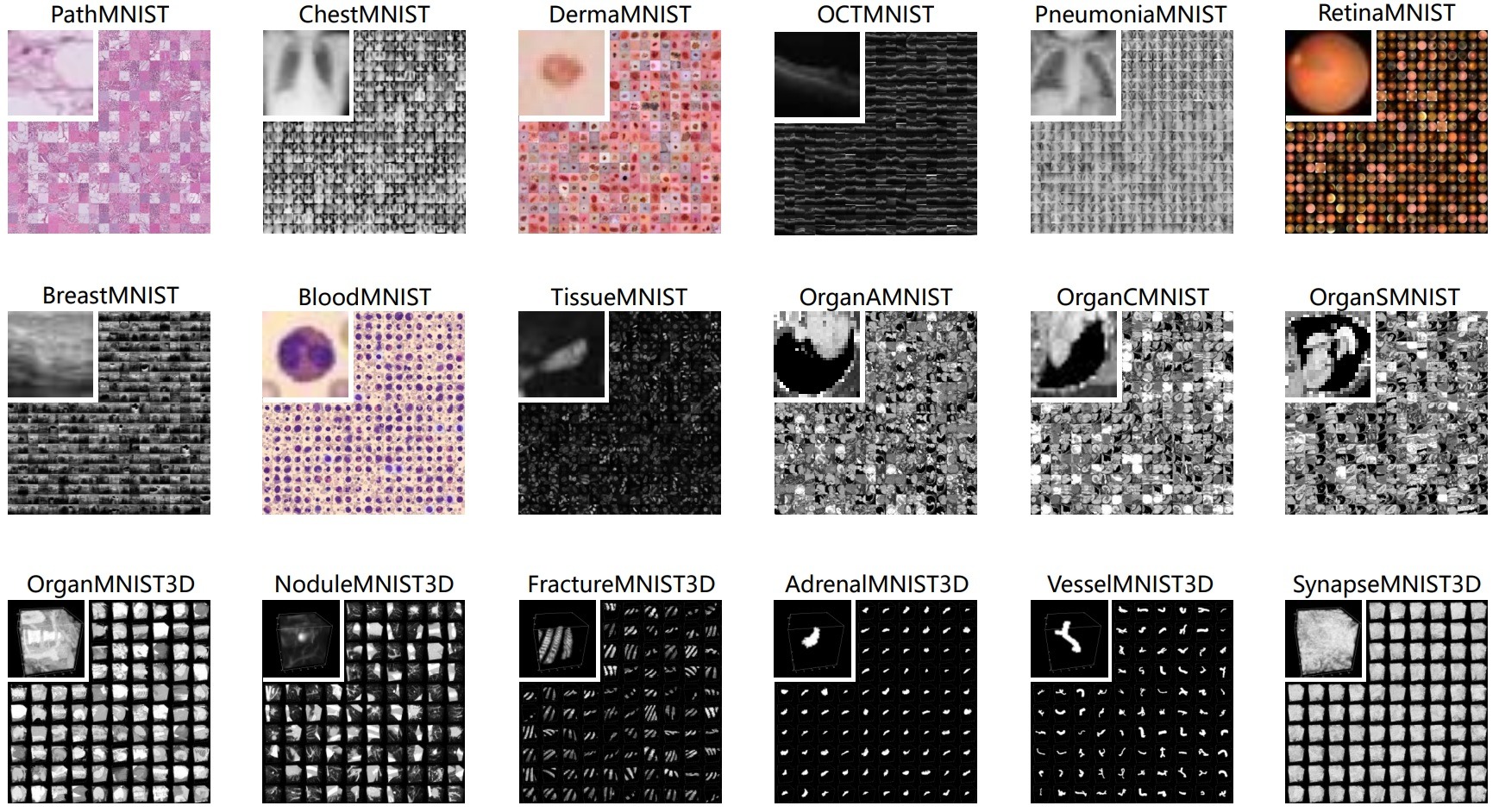}
	\caption{Overview of the MedMNIST v2 dataset}
	\label{images:medmnist_v2}
\end{figure*}

Yang et al. ~\citep{medmnistv2} presented the overview of the MedMNIST v2 dataset as Figure \ref{images:medmnist_v2} shown.

\subsubsection{Experimental Setup}
Again, we're using the models in the Table ~\ref{tab:overall-architecture}, and we're using a configuration called 3d, which might as well be thought of as a 3D version of LeNet. Considering that the size of the 3D datasets used in the experiment is the same, the network applied to different types of datasets is practically the same. For training details, we refer to the setup of the MedMnistv2 paper~\citep{medmnistv2}. They utilize an Adam optimizer with an initial learning rate of 0.001 and train the model for 100 epochs, delaying the learning rate by 0.1 after 50 and 75 epochs. Additionally, as a regularization for the two datasets of shape modality (i.e., AdrenalMNIST3D/VesselMNIST3D). For the batch size, We set the size of 28 $\times$ 28 $\times$ 28 to 32, which is the same as the original paper; Limited by our computing power, on a 64 $\times$  64 $\times$  64 dimension, we set the batch size to 2. We repeat the experiment 5 times for each model and report the mean and variance of the experimental metrics.

As mentioned earlier, on the same size of the 3D dataset, such as 28 $\times$ 28 $\times$ 28 or 64 $\times$  64 $\times$  64 , in fact, the architecture of the model is exactly the same except for the classifier, such as the feature extraction network, so in order to compare the generalization ability of the model, we also refer to the original text and calculate the average of the indicators of the same model on each dataset to further compare the performance of the model.

\subsubsection{Results}
The Table \ref{tab:comparison_MedMNIST28_results} compares various LeNet-based methods on several $28 \times 28 \times 28$ MNIST3D datasets, including OrganMNIST3D, NoduleMNIST3D, AdrenalMNIST3D, FractureMNIST3D, VesselMNIST3D, and SynapseMNIST3D. The performance metrics evaluated are Area Under the Curve (AUC) and Accuracy (ACC). The results reveal that different variants of the LeNet architecture exhibit varying degrees of effectiveness across these $28 \times 28 \times 28$ datasets.Among the methods, LeNet-C-$\alpha / \beta$(28) consistently performs well, especially on the OrganMNIST3D and FractureMNIST3D datasets, where it achieves an AUC of 0.987 and 0.640, and ACC values of 87.213 and 48.750, respectively and it even outperforms the benchmark on the AdrenalMNIST3D dataset. This indicates that incorporating compound tropical convolution mechanisms (denoted by "C"  in the notation) significantly enhances the model's ability to generalize across diverse medical imaging tasks. Notably, LeNet-P-$\alpha / \beta$(28) also shows strong performance, suggesting that the parallel tropical convolution method is effective even without the additional modifications present in the P variants.The traditional LeNet(28) serves as a baseline for comparison, and while it achieves acceptable performance in some datasets, such as VesselMNIST3D with an AUC of 0.495 and ACC of 88.743, its performance on other datasets, like OrganMNIST3D and FractureMNIST3D, is suboptimal with ACC values as low as 25.115 and 37.500, respectively. This highlights the limitations of the original architecture when applied to more complex 3D medical imaging tasks.Comparatively, methods like LeNet-CM-$\alpha / \beta$(28) and LeNet-F-$\Rmnum{3}$(28) show competitive performance, particularly in datasets like AdrenalMNIST3D and SynapseMNIST3D, demonstrating the effectiveness of convolutional modifications and feature integration strategies in enhancing model performance. However, their overall superiority is less consistent across all datasets when compared to the PM variants.

The Table \ref{tab:comparison_MedMNIST64_results} provides a comparative analysis of several LeNet-based methods with a larger input size ($64 \times 64 \times 64$) on various MNIST3D datasets, including OrganMNIST3D, NoduleMNIST3D, AdrenalMNIST3D, FractureMNIST3D, VesselMNIST3D, and SynapseMNIST3D. The performance metrics considered are Area Under the Curve (AUC) and Accuracy (ACC). This analysis allows us to assess how these methods scale with increased input resolution and whether they maintain or improve performance on complex 3D medical imaging tasks. The performance of the LeNet architecture with $64\times64\times64$ input size varies significantly across different methods and datasets. Notably, LeNet-F-\Rmnum{2}(64) achieves high AUC values, particularly on the OrganMNIST3D and NoduleMNIST3D datasets, with AUCs of 0.967 and 0.840, respectively. Its ACC values on these datasets are also commendable, reaching 80.131\% and 83.613\%, respectively. This indicates that the feature integration approach used in the F-\Rmnum{2} variant is effective for these specific datasets when the input resolution is increased. However, other variants, such as LeNet-PM-$\alpha / \beta$(64), which performed well in the previous experiment with $28\times28\times28$ input size, show a notable decline in performance with the $64\times64\times64$ input size. For example, on the OrganMNIST3D dataset, this method records an AUC of only 0.500 and an ACC of 11.279\%, suggesting that the model struggles to generalize when the input size increases. Similar declines are observed across other datasets for this variant, indicating that certain architectural modifications that performed well at lower resolutions may not scale effectively. The original LeNet(64) also demonstrates varying performance. While it achieves reasonable results on the AdrenalMNIST3D and VesselMNIST3D datasets, with ACC values of 76.846\% and 88.743\%, respectively, its performance on other datasets such as OrganMNIST3D is considerably lower, with an AUC of 0.500 and an ACC of 11.279\%. This highlights the challenges faced by the original architecture when processing higher-resolution inputs in more complex tasks. Among the convolutional modifications, LeNet-C-$\alpha / \beta$(64) shows some promise, particularly on the AdrenalMNIST3D dataset with an AUC of 0.705 and an ACC of 80.537\%, indicating that the incorporation of compound tropical convolutional modifications helps improve performance in specific cases. However, this method does not consistently outperform others across all datasets, suggesting that the effectiveness of compound tropical convolutional modification may be dataset-dependent.

The experimental results underscore the benefits of architectural modifications to the LeNet network, highlighting how these changes contribute to performance improvements across different input resolutions. For the $28\times28\times28$ input size, the introduction of parallel tropical convolutional modifications, as seen in variants like LeNet-PM-$\alpha / \beta$(28), significantly enhances performance across multiple datasets. These modifications allow the network to better capture and integrate complex features, resulting in superior AUC and ACC metrics on challenging tasks like OrganMNIST3D and VesselMNIST3D. The parallel tropical convolution method, in particular, proves to be highly effective in boosting the model's ability to generalize across diverse medical imaging tasks, showcasing the advantage of fine-tuning the network's architecture for low-resolution inputs.When the input size is increased to $64\times64\times64$, certain architectural changes continue to demonstrate their effectiveness, albeit with some variations. LeNet-F-\Rmnum{2}(64), which integrates advanced feature processing techniques, performs exceptionally well on datasets like OrganMNIST3D and NoduleMNIST3D. This suggests that the feature integration strategies employed in this variant scale effectively with higher-resolution inputs, enabling the network to maintain strong performance by capturing more detailed information from the larger input dimensions. However, it is also evident that some modifications, particularly those in LeNet-PM-$\alpha / \beta$(64), do not translate as effectively to higher resolutions, indicating that not all enhancements are universally beneficial across different input scales.At the same time, it is worth noting that the architectural modification based on tropical convolution can also reduce the amount of computation required by the model to a certain extent, and achieve a better computational-performance balance for the model in medical image classification tasks.

In summary, the architectural modifications to the LeNet network provide clear advantages at both input resolutions. For $28\times28\times28$ inputs, modifications like parallel tropical convolution and compound tropical convolution are particularly beneficial, offering substantial gains in performance. For $64\times64\times64$ inputs, feature integration methods such as those in LeNet-F-\Rmnum{2}(64) demonstrate their value, effectively leveraging the increased input size to capture more complex patterns. These results highlight models based on tropical convolutions can bring better nonlinear fitting ability, which leads to better model performance, while reducing the dependence on computing resources

\begin{table*}[htbp]
	\centering
	\caption{Comparison of various LeNet-based architectures on six $28\times 28 \times 28$ MNIST3D datasets: OrganMNIST3D, NoduleMNIST3D, AdrenalMNIST3D, FractureMNIST3D, VesselMNIST3D, and SynapseMNIST3D. The performance is evaluated using Area Under the Curve (AUC) and Accuracy (ACC). The models include standard LeNet, LeNet with ReLU, and various LeNet variants based tropical convolution. The benchmark scores for each dataset are also provided.}
	\label{tab:comparison_MedMNIST28_results}
	\resizebox{\textwidth}{!}{
	\begin{tabular}{|l|cc|cc|cc|cc|cc|cc|}
		\toprule
		Methods
		& \multicolumn{2}{c|}{OrganMNIST3D} & \multicolumn{2}{c|}{NoduleMNIST3D} & \multicolumn{2}{c|}{AdrenalMNIST3D} & \multicolumn{2}{c|}{FractureMNIST3D} & \multicolumn{2}{c|}{VesselMNIST3D} & \multicolumn{2}{c|}{SynapseMNIST3D} \\
		\cline{2-13}
		& AUC & ACC & AUC & ACC & AUC & ACC & AUC & ACC & AUC & ACC & AUC & ACC \\
		\midrule
LeNet(28) & 0.629 & 25.115              & 0.618 & 79.355              & 0.537 & 76.846              & 0.527 & 37.500              & 0.495 & 88.743              & 0.508 & 73.011  \\ 
LeNet-ReLU(28) & 0.976 & 80.656              & 0.839 & 83.484              & 0.786 & 80.671              & 0.579 & 40.333              & 0.918 & \textbf{92.618 }             & 0.614 & 71.364  \\ 
LeNet-F-\Rmnum{1}(28) & 0.962 & 77.803              & 0.825 & 82.387              & 0.827 & 81.409              & 0.623 & 44.250              & \textbf{0.938} & 92.565              & 0.713 & 72.841  \\ 
LeNet-F-\Rmnum{2}(28) & 0.965 & 77.967              & 0.830 & 81.613              & \textbf{0.865} & 82.215              & 0.545 & 41.667              & 0.909 & 91.414              & 0.708 & 73.182  \\ 
LeNet-F-\Rmnum{3}(28) & 0.966 & 79.180              & 0.796 & 79.226              & 0.861 & 82.953              & 0.610 & 43.917              & 0.888 & 90.314              & 0.697 & 72.614  \\ 
LeNet-C-$\alpha$(28) & 0.983 & 84.426              & 0.881 & 84.645              & 0.840 & 79.799              & 0.607 & 45.250              & 0.847 & 89.424              & 0.783 &\textbf{ 78.352}  \\ 
LeNet-C-$\alpha / \beta$(28) & \textbf{0.987} & \textbf{87.213}              & \textbf{0.887} & 84.387              & 0.859 & \textbf{83.221 }             & \textbf{0.640 }& \textbf{48.750}              & 0.868 & 90.105              & 0.779 & 77.102  \\ 
LeNet-CM-$\alpha$(28) & 0.974 & 81.213              & 0.867 & 83.613              & 0.763 & 78.658              & 0.549 & 41.833              & 0.867 & 90.157              & 0.741 & 75.511  \\ 
LeNet-CM-$\alpha / \beta$(28) & 0.978 & 82.426              & 0.870 & 84.258              & 0.753 & 77.450              & 0.561 & 42.917              & 0.877 & 90.419              & 0.757 & 76.648  \\ 
LeNet-P-$\alpha$(28) & 0.984 & 84.426              & 0.880 & \textbf{84.839}              & 0.847 & 80.403              & 0.614 & 46.917              & 0.853 & 90.000              & 0.786 & 78.068  \\ 
LeNet-P-$\alpha / \beta$(28) & 0.986 & 85.770              & 0.886 & 84.452              & 0.837 & 80.470              & 0.634 & 47.667              & 0.886 & 90.942              & \textbf{0.796} & 77.727  \\ 
LeNet-PM-$\alpha$(28) & 0.976 & 81.836              & 0.868 & 83.742              & 0.763 & 79.128              & 0.564 & 43.250              & 0.878 & 90.419              & 0.735 & 75.852  \\ 
LeNet-PM-$\alpha / \beta$(28) & 0.977 & 82.459              & 0.864 & 83.613              & 0.761 & 77.651              & 0.553 & 43.000              & 0.866 & 89.738              & 0.746 & 77.500  \\ 
\hline
		Benchmark & 0.996 & 90.700 & 0.914 & 87.400 & 0.839 & 80.200 & 0.754 & 51.700 & 0.930 & 92.800 & 0.851 & 79.500 \\
		\bottomrule
	\end{tabular}
}

\end{table*}

\begin{table*}[htbp]
	\centering
	\caption{Comparison of various LeNet-based architectures on six $64\times 64 \times 64$ MNIST3D datasets: OrganMNIST3D, NoduleMNIST3D, AdrenalMNIST3D, FractureMNIST3D, VesselMNIST3D, and SynapseMNIST3D. The performance is evaluated using Area Under the Curve (AUC) and Accuracy (ACC). The models include standard LeNet, LeNet with ReLU, and various LeNet variants based tropical convolution. The benchmark scores for each dataset are also provided.}
	\label{tab:comparison_MedMNIST64_results}
	\resizebox{\textwidth}{!}{
	\begin{tabular}{|l|cc|cc|cc|cc|cc|cc|}
		\toprule
		Methods & \multicolumn{2}{c|}{OrganMNIST3D} & \multicolumn{2}{c|}{NoduleMNIST3D} & \multicolumn{2}{c|}{AdrenalMNIST3D} & \multicolumn{2}{c|}{FractureMNIST3D} & \multicolumn{2}{c|}{VesselMNIST3D} & \multicolumn{2}{c|}{SynapseMNIST3D} \\
		\cline{2-13}
		& AUC & ACC & AUC & ACC & AUC & ACC & AUC & ACC & AUC & ACC & AUC & ACC \\
		\midrule
LeNet(64) & 0.500 & 11.279              & 0.500 & 79.355              & 0.500 & 76.846              & 0.500 & 37.500              & 0.500 & 88.743              & 0.500 & 73.011  \\ 
LeNet-ReLU(64) & 0.951 & 71.672              & 0.617 & 80.387              & \textbf{0.771} & 80.201              & 0.551 & 41.500              & 0.554 & 89.110              & 0.508 & 72.557  \\ 
LeNet-F-\Rmnum{1}(64) & 0.693 & 25.607              & 0.656 & 80.710              & 0.674 & 78.188              & \textbf{0.651} & 46.833              & 0.829 & 90.314              & 0.500 & 73.011  \\ 
LeNet-F-\Rmnum{2}(64) & \textbf{0.967} & \textbf{80.131}              & \textbf{0.840} & 83.613              & 0.718 & 77.785              & 0.639 & \textbf{48.083}              & \textbf{0.835} & 75.969              & 0.605 & 68.580  \\ 
LeNet-F-\Rmnum{3}(64) & 0.847 & 58.492              & 0.821 & \textbf{83.677}              & 0.579 & 78.188              & 0.594 & 43.667              & 0.572 & 89.319              & 0.634 & 71.136  \\ 
LeNet-C-$\alpha$(64) & 0.865 & 63.836              & 0.569 & 68.387              & 0.548 & 77.584              & 0.525 & 39.083              & 0.548 & 89.424              & 0.500 & 73.011  \\ 
LeNet-C-$\alpha / \beta$(64) & 0.951 & 74.918              & 0.577 & 80.710              & 0.705 & \textbf{80.537}              & 0.586 & 43.417              & 0.450 & 88.743              & 0.500 & 63.807  \\ 
LeNet-CM-$\alpha$(64) & 0.767 & 48.557              & 0.707 & 82.129              & 0.627 & 68.255              & 0.535 & 40.250              & 0.648 & 90.105              & 0.500 & 73.011  \\ 
LeNet-CM-$\alpha / \beta$(64) & 0.682 & 36.918              & 0.716 & 82.323              & 0.679 & 68.255              & 0.545 & 39.833              & 0.637 & 89.895              & 0.500 & 73.011  \\ 
LeNet-P-$\alpha$(64) & 0.953 & 75.836              & 0.570 & 56.065              & 0.571 & 77.651              & 0.558 & 42.333              & 0.461 & 88.743              & 0.500 & 73.011  \\ 
LeNet-P-$\alpha / \beta$(64) & 0.883 & 68.262              & 0.725 & 71.226              & 0.638 & 79.732              & 0.590 & 44.833              & 0.544 & 89.424              & 0.500 & 73.011  \\ 
LeNet-PM-$\alpha$(64) & 0.774 & 50.230              & 0.763 & 82.000              & 0.679 & 79.799              & 0.560 & 42.083              & 0.708 & \textbf{90.419}              & 0.529 & 64.205  \\ 
LeNet-PM-$\alpha / \beta$(64) & 0.500 & 11.279              & 0.696 & 82.194              & 0.725 & 80.268              & 0.571 & 42.583              & 0.567 & 88.796              & 0.500 & 73.011  \\ 

\hline
	Benchmark & 0.996 & 90.700 & 0.914 & 87.400 & 0.839 & 80.200 & 0.754 & 51.700 & 0.930 & 92.800 & 0.851 & 79.500 \\
		\bottomrule
	\end{tabular}
}

\end{table*}

We compare the performance of the model by calculating the average of the model's evaluation metrics on various datasets, as shown in Table ~\ref{tab:average_performance_medmnist28} and Table ~\ref{tab:average_performance_medmnist64}.

The experimental results summarized in Tables  ~\ref{tab:average_performance_medmnist28} and ~\ref{tab:average_performance_medmnist64} highlight the performance disparities of various LeNet-based architectures across MedMNISTv2 datasets with different input dimensions. On the smaller $28\times28\times28$ datasets, the LeNet-P-$\alpha / \beta$(28) and LeNet-C-$\alpha / \beta$(28) architectures exhibit superior performance, achieving the highest average AUC and accuracy, with LeNet-P-$\alpha / \beta$(28) leading at an average AUC of 0.838 and an average accuracy of 77.838\%  and LeNet-C-$\alpha / \beta$(28) leading at an average AUC of 0.836 and an average accuracy of 78.463\%. These results underscore the effectiveness of the $\alpha / \beta$ convolutional modifications in handling smaller input sizes. However, when applied to the larger 64×64×64 datasets, the performance of these architectures notably declines, with no model achieving comparable results to those observed on the smaller datasets. LeNet-F-\Rmnum{2}(64) performs best among the deeper models on larger inputs but still falls short of the effectiveness seen with $28\times28\times28$ inputs, reflecting an overall challenge in scaling these architectures. These findings suggest that while the $\alpha / \beta$ variants enhance performance for smaller inputs, they struggle to maintain their efficacy with increased input dimensionality. By calculating the average data, we can see that while the tropical convolutional method reduces the amount of computation brought about by its own design, it can still obtain similar or even better performance than the standard convolution model, reflecting a more reasonable computation-performance balance.

\begin{table}[htbp]
	\centering
	\small
	\setlength{\tabcolsep}{3pt}
	\caption{Average performance of MedMNISTv2 in metrics of average AUC and average ACC over all 3D 28 $\times$ 28 $\times$ 28 datasets.}
	\label{tab:average_performance_medmnist28}
	\begin{tabular}{|l|c|c|}
		\hline
		\textbf{Methods} & \textbf{AVG AUC} & \textbf{AVG ACC} \\ \hline
LeNet(28) & 0.552 & 63.428  \\  \hline
LeNet-ReLU(28) & 0.785 & 74.854  \\  \hline
LeNet-F-\Rmnum{1}(28) & 0.814 & 75.209  \\  \hline
LeNet-F-\Rmnum{2}(28) & 0.804 & 74.676  \\  \hline
LeNet-F-\Rmnum{3}(28) & 0.803 & 74.701  \\  \hline
LeNet-C-$\alpha$(28) & 0.823 & 76.983  \\  \hline
LeNet-C-$\alpha / \beta$(28) & 0.836 & \textbf{78.463}  \\  \hline
LeNet-CM-$\alpha$(28) & 0.794 & 75.164  \\  \hline
LeNet-CM-$\alpha / \beta$(28) & 0.799 & 75.686  \\  \hline
LeNet-P-$\alpha$(28) & 0.827 & 77.442  \\  \hline
LeNet-P-$\alpha / \beta$(28) & \textbf{0.838} & 77.838  \\  \hline
LeNet-PM-$\alpha$(28) & 0.797 & 75.704  \\  \hline
LeNet-PM-$\alpha / \beta$(28) & 0.794 & 75.660  \\  \hline

	\end{tabular}
    
\end{table}

\begin{table}[htbp]
	\centering
	\small
	\setlength{\tabcolsep}{3pt}
	\caption{Average performance of MedMNISTv2 in metrics of average AUC and average ACC over all 3D 64 $\times$ 64 $\times$ 64  datasets.}
	\label{tab:average_performance_medmnist64}
	\begin{tabular}{|l|c|c|}
		\hline
		\textbf{Methods} & \textbf{AVG AUC} & \textbf{AVG ACC} \\ \hline
LeNet(64) & 0.500 & 61.122  \\  \hline
LeNet-ReLU(64) & 0.659 & \textbf{72.571}  \\  \hline
LeNet-F-\Rmnum{1}(64) & 0.667 & 65.777  \\  \hline
LeNet-F-\Rmnum{2}(64) & \textbf{0.767} & 72.360  \\  \hline
LeNet-F-\Rmnum{3}(64) & 0.675 & 70.747  \\  \hline
LeNet-C-$\alpha$(64) & 0.592 & 68.554  \\  \hline
LeNet-C-$\alpha / \beta$(64) & 0.628 & 72.022  \\  \hline
LeNet-CM-$\alpha$(64) & 0.631 & 67.051  \\  \hline
LeNet-CM-$\alpha / \beta$(64) & 0.627 & 65.039  \\  \hline
LeNet-P-$\alpha$(64) & 0.602 & 68.940  \\  \hline
LeNet-P-$\alpha / \beta$(64) & 0.647 & 71.081  \\  \hline
LeNet-PM-$\alpha$(64) & 0.669 & 68.122  \\  \hline
LeNet-PM-$\alpha / \beta$(64) & 0.593 & 63.022  \\  \hline

	\end{tabular}
    
\end{table}

The comprehensive evaluation of LeNet-based architectures across various MedMNISTv2 datasets, as detailed in Tables ~\ref{tab:comparison_MedMNIST28_results}, ~\ref{tab:comparison_MedMNIST64_results}, ~\ref{tab:comparison_MedMNIST28_results}, and ~\ref{tab:average_performance_medmnist64}, reveals several key insights. Notably, the $\alpha / \beta$ convolutional modifications consistently enhance model performance, particularly on smaller $28\times28\times28$ input datasets, where LeNet-P-$\alpha / \beta$(28) achieved the highest average AUC of 0.838 and accuracy of 77.838\%. However, as input size increases to $64\times64\times64$, a significant decline in performance is observed across all models, indicating that depth alone does not compensate for the challenges posed by larger datasets. Despite some success on specific datasets like OrganMNIST3D and NoduleMNIST3D, the LeNet variants struggle with more complex datasets, such as FractureMNIST3D and SynapseMNIST3D, where even the best models achieved relatively low accuracy. These findings suggest that while targeted architectural modifications can yield substantial gains, further innovation is needed to develop models that can consistently perform well across varying dataset sizes and complexities.

\subsection{Exploring Efficient Yet Simplified Structures}
We noticed that no matter compound tropical convolutional layers or parallel tropical convolutional layers, they introduced an additional parameter, such as only $\alpha$ or $\alpha$ and $\beta$, in the process of combining feature maps. We were curious whether these two parameters were necessary, so we considered that the direct summation of the feature values on the two convolution branches was not a learnable weighted paramter summation. We implemented and named the four network variants corresponding to these four modes LeNet-CC, LeNet-CCM, LeNet-CP, and LeNet-CPM respectively; corresponding to the LeNet-C-$\alpha$ or LeNet-C-$\alpha/\beta$, LeNet-CM-$\alpha$ or LeNet-CM-$\alpha/\beta$, LeNet-P-$\alpha$ or LeNet-P-$\alpha/\beta$, and LeNet-PM-$\alpha$ or LeNet-PM-$\alpha/\beta$ models in the previous section.

We use the same experimental settings as in the previous sections and train/test these variants on the one-dimensional and two-dimensional datasets. The results are summarized in Table~\ref{tab:constan_model_performance}.

\begin{table*}[!t]
\centering
\caption{Comparison of Simplified Structures Model Performance Across 1D and 2D Datasets}
\begin{tabular}{|l|l|r|c|c|}
\hline
\textbf{Dataset} & \textbf{Model} & \textbf{Parameters (M)} & \textbf{Accuracy (\%)} & \textbf{AUC} \\
\hline
\multirow{4}{*}{UrbanSound8K} & LeNet-CP   & 14.62392 & 40.10766 $ \pm $  21.03064 &  0.74301 $ \pm $0.14147 \\
							  & LeNet-CPM  & 14.62365 & 88.05573 $ \pm $  0.88174 &0.98076  $ \pm $ 0.00343  \\
& LeNet-CC   & 14.62319 & 20.40532 $ \pm $ 6.79555& 0.61494 $ \pm $ 0.08684  \\
							  & LeNet-CCM  & 14.62319 & 71.65294 $ \pm $ 30.31767 & 0.87995 $ \pm $ 0.18998  \\

\hline
\multirow{4}{*}{\begin{tabular}{@{}l@{}}Speech \\ Command\end{tabular}} 
							  & LeNet-CP   & 3.63962 & 17.14857$ \pm $4.85247 & 0.77936$ \pm $0.05140 \\
							  & LeNet-CPM  & 3.63934 & 44.61427$ \pm $0.15133 & 0.91680$ \pm $0.00088 \\
							  & LeNet-CC   & 3.63889 & 16.39618$ \pm $3.76647 & 0.77514$ \pm $0.03910  \\
							  & LeNet-CCM  & 3.63889 & 44.29986$ \pm $1.36773 & 0.92180$ \pm $0.00428  \\
\hline
\multirow{4}{*}{\begin{tabular}{@{}l@{}}ECG \\ Heartbeat \\ Categorization \end{tabular}} 
							  & LeNet-CP   & 0.06114 & 97.60460 $ \pm $ 0.27233& 0.98127$ \pm $0.00187 \\
							  & LeNet-CPM  & 0.06086 & 98.10616 $ \pm $ 0.16110& 0.98448$ \pm $0.00135 \\
							  & LeNet-CC   & 0.06040 & 97.40819 $ \pm $ 0.35023& 0.98131$ \pm $0.00084 \\
							  & LeNet-CCM  & 0.06041 & 98.15914 $ \pm $ 0.07875& 0.98515$ \pm $0.00052 \\

\hline
\hline
\multirow{4}{*}{MNIST} & LeNet-CP   & 0.06128 & 98.92200$ \pm $ 0.06997& 0.99991$ \pm $ 0.00003\\
					   & LeNet-CPM  & 0.05899 & 99.11600$ \pm $ 0.09972& 0.99992$ \pm $ 0.00003\\
					   & LeNet-CC   & 0.05885 & 98.90600$ \pm $ 0.22402& 0.99989$ \pm $ 0.00004 \\
					   & LeNet-CCM  & 0.05885 & 98.92200$ \pm $ 0.21885& 0.99989$ \pm $ 0.00008\\
\hline
\multirow{4}{*}{\begin{tabular}{@{}c@{}}Fashion \\ MNIST\end{tabular}} 
							  & LeNet-CP   & 0.06128 & 89.07000$ \pm $0.65568 & 0.99159 $ \pm $0.00092 \\
							  & LeNet-CPM  & 0.05899 & 90.87400$ \pm $0.57964 & 0.99366$ \pm $ 0.00059\\
							  & LeNet-CC   & 0.05885 & 88.60800$ \pm $0.55787 & 0.99107$ \pm $ 0.00076\\
							  & LeNet-CCM  & 0.05885 & 91.02400$ \pm $0.73123 & 0.99374$ \pm $ 0.00077 \\
\hline
\multirow{4}{*}{SVHN}         & LeNet-CP   & 0.08199 & 80.99877$ \pm $8.29775 & 0.96926$ \pm $ 0.02677\\
							  & LeNet-CPM  & 0.07970 & 87.33482$ \pm $0.78928 & 0.98619$ \pm $ 0.00139 \\
							  & LeNet-CC   & 0.07928 & 81.14551$ \pm $7.85800 & 0.97053$ \pm $ 0.02443\\
							  & LeNet-CCM  & 0.07928 & 87.49462$ \pm $0.49121 & 0.98679$ \pm $ 0.00075 \\
\hline
\multirow{4}{*}{CIFAR-10}     & LeNet-CP   & 0.08199 & 53.74200$ \pm $ 3.22195&0.89620$ \pm $0.01236 \\
							  & LeNet-CPM  & 0.07970 & 64.06400$ \pm $ 2.08353&0.93440$ \pm $0.00663 \\
							  & LeNet-CC   & 0.07928 & 53.08400 $ \pm$ 2.76169&0.89305$ \pm $0.01057 \\
							  & LeNet-CCM  & 0.07928 & 63.83400$ \pm $ 2.18436&0.93360$ \pm $0.00687 \\
\hline
\end{tabular}
\label{tab:constan_model_performance}
\end{table*}

Based on the experimental results presented in Table \ref{tab:1d-TCNN-results} and Table \ref{tab:constan_model_performance}, we can observe the performance comparison between the original models and their simplified versions. In Table \ref{tab:1d-TCNN-results}, various models, such as LeNet-P, LeNet-PM, LeNet-C, and LeNet-CM, are tested across three different datasets: UrbanSound8K, Speech Command, and ECG Heartbeat Categorization. These models show promising results in terms of accuracy, AUC (Area Under the Curve), F1 score, precision, and recall, while also considering the number of operations and model parameters. For instance, LeNet-CM achieved an accuracy of 87\% on the UrbanSound8K dataset.

In contrast, Table \ref{tab:constan_model_performance} shows the simplified versions of these models—LeNet-CP, LeNet-CPM,LeNet-CC, and LeNet-CC designed to reduce computational complexity while maintaining competitive performance. Despite the reduction in parameters and operations, the simplified models continue to perform effectively across all datasets. For example, LeNet-CPM achieved an accuracy of 88.05\% on the UrbanSound8K dataset, which is even slightly higher than the original model LeNet-CM and very competitive. Moreover, the simplifications resulted in significantly fewer parameters, demonstrating the potential for deployment in resource-constrained environments. The simplified LeNet-CC model also showed an impressive AUC value of 0.98131 on the ECG Heartbeat Categorization dataset, indicating its strong performance in clinical applications despite the reduced complexity.

As seen in Table \ref{tab:2dc-TCNN-results}, the full models achieved high accuracy and AUC across all datasets, with the MNIST dataset showing accuracy up to 98.96\% (LeNet-P-$\alpha / \beta$) and AUC values of 1.00 consistently across most configurations. Similarly, Fashion MNIST and SVHN achieved comparable performance, with the top configurations reaching accuracies of 91.11\% and 88.86\%, respectively. On CIFAR-10, the performance was slightly lower due to the increased complexity of the dataset, with the best accuracy at 66.82\% (LeNet-P-$\alpha / \beta$).

The simplified models, as illustrated in Table \ref{tab:constan_model_performance}, demonstrated a significant reduction in the number of parameters while retaining competitive accuracy and AUC. For MNIST, the simplified LeNet-CPM model achieved 99.116\% accuracy with only 0.05899M parameters. Similar trends were observed across other datasets. On Fashion MNIST, the LeNet-CCM model achieved an accuracy of 91.024\% with 0.05885M parameters, showcasing its ability to balance efficiency and accuracy. For SVHN, the simplified models reached an accuracy of 87.49\%, which is comparable to the full models, with a significant reduction in complexity. On the more challenging CIFAR-10 dataset, the simplified models maintained reasonable accuracy (up to 63.83\%) while requiring only 0.07928M parameters, demonstrating their applicability in scenarios where computational resources are limited.

These results highlight the trade-offs between model complexity and performance. The simplified models, despite their reduced parameter counts, were able to maintain a competitive level of accuracy and AUC across multiple datasets. This indicates their potential for deployment in resource-constrained environments without significant loss of performance. Moreover, the simplified architectures exhibited strong generalization across diverse datasets, further validating their robustness and applicability in real-world scenarios.

\FloatBarrier


{\small
\bibliographystyle{ieee}
\bibliography{Ref}
}

\end{document}